# RoboSolver: A Multi-Agent Large Language Model Framework for Solving Robotic Arm Problems

Hamid Khabazi[1], Ali F. Meghdari[1,2], Alireza Taheri[1,3,*]

[1]Social & Cognitive Robotics Laboratory, Center of Excellence in Design, Robotics, and Automation (CEDRA), Sharif University of Technology, Tehran, Iran

[2] Chancellor, Fereshtegaan International Branch, Islamic Azad University, Tehran, Iran

[3] Artificial Intelligence in Design and Complex Systems (AIDACS) Group, Sharif University of Technology, Tehran, Iran

[*]**Corresponding Author: artaheri@sharif.edu , Tel: +982166165531**

**Abstract**

One major challenge for engineering students learning advanced topics—especially through question-and-answer interactions with LLMs or VLMs—is that fields such as robotics are fundamentally conceptual rather than memorization-based. Mastery of this subject requires spatial reasoning and mathematical skills that LLMs or VLMs alone cannot fully support. To address these limitations, this study proposes an intelligent multi-agent framework built on LLMs and VLMs and specifically tailored to robotics. The goal is to integrate the strengths of LLMs and VLMs with computational tools to automatically analyze and solve problems related to robotic manipulators. Our developed framework accepts both textual and visual inputs and can automatically perform forward and inverse kinematics, compute velocities and accelerations of key points, generate 3D simulations of the robot, and ultimately execute motion control within the simulated environment, all according to the user's query. To evaluate the framework, three benchmark tests were designed, each consisting of ten questions. In the first benchmark test, the framework was evaluated while connected to GPT-4o, DeepSeek-V3.2, and Claude-Sonnet-4.5, as well as their corresponding raw models. The objective was to extract the forward kinematics of robots directly from textual descriptions. The results showed that the framework integrated with GPT-4o achieved the highest accuracy, reaching 0.97 in computing the final solution, whereas the raw model alone attained an accuracy of only 0.30 for the same task. Similarly, for the other two models, the framework consistently outperformed the corresponding raw models in terms of accuracy. The second benchmark test was identical to the first, except that the input was provided in visual form. In this test, the GPT-4o LLM was used alongside the Gemini 2.5 Pro VLM. The results showed that the framework achieved an accuracy of 0.93 in obtaining the final answer, which is approximately 20% higher than that of the corresponding raw model. The third benchmark test encompassed a range of robotic tasks, including simulation, control, velocity and acceleration computation, as well as inverse kinematics and Jacobian calculation, for which the framework achieved an accuracy of 0.97.

# 1 INTRODUCTION

In recent years, Artificial Intelligence (AI) has experienced widespread expansion across various fields, leading to a fundamental transformation in diverse application domains. With advances in Transformer architectures [1], Large Language Models (LLMs) have emerged, enhancing human–machine interaction and opening new research avenues for autonomous systems capable of human-like reasoning. For example, Yiwei Wang et al. proposed an LLM-based autonomous agent for a robotic system, enabling effective human–robot collaboration and allowing the robot to understand natural language instructions and execute complex industrial tasks autonomously [2]. In another study, Shuguang Liu et al. [3] investigated the problem of motion planning for industrial mobile robots in factory environments with the assistance of LLMs. Furthermore, at a lower level of robotic operation, LLMs have also been utilized in the adaptive control of robotic manipulators [4]. Moreover, LLMs are not limited to industrial robotics applications. In this regard, Merino-Fidalgo et al. employed an LLM as the decision-making core of the social robot Temi, enabling it to carry out tasks such as verbal interaction, reminders, and autonomous navigation within a domestic environment [5]. Overall, the emergence of LLMs has enabled the automation of numerous tasks that previously required substantial human involvement. Furthermore, as mentioned, LLMs possess strong natural language reasoning capabilities. Consequently, one of their most common applications is quick and effortless access to information through question answering. For instance, individuals can pose their inquiries to these models and, in many cases, receive highly appropriate and reliable responses. Therefore, the capacity for serious utilization of these models in the field of education has expanded significantly in recent years. For example, Latif et al. introduced a bot called PhysicsAssistant which, by leveraging GPT-3.5-turbo, plays the role of an intelligent educational assistant in a physics laboratory setting, assisting students with laboratory activities to facilitate the learning process [6]. Furthermore, OpenAI has recently added a new feature called Study Mode to ChatGPT. This feature is designed to support the learning process of students. In this mode, instead of directly providing the correct answer, the language model first assesses the user's level of understanding and knowledge by posing relevant questions. Subsequently, much like a tutor, it guides the user step-by-step, encouraging active participation, critical thinking, and engagement with the problem, which ultimately leads to deeper user learning [7]. In addition to these studies, Bo Ni et al. introduced a platform supported by multi-agents based on language models that has the capability to automatically solve elasticity problems using the Finite Element Method (FEM) [8]. This type of utilization of language models for solving scientific problems requires the precise design of interactions and the purposeful guidance of the model towards a solution. To achieve such goals, Prompt Engineering and intelligent agents based on language models are employed. Prompt Engineering is the process of designing precise and effective instructions, known as prompts, with the goal of directing the language models towards producing desired outputs. An agent is also a language model combined with auxiliary tools to extend its capabilities. For instance, since language models are trained solely on massive volumes of text, they perform simple operations like addition and subtraction based only on the textual patterns they hold in memory, lacking genuine mathematical understanding. However, by converting the language model into an intelligent agent, it can be equipped with tools such as a calculator. In this scenario, when a mathematical question is asked, instead of attempting to answer based solely on pre-learned data (and guessing the next token), it sends the problem parameters to the calculator function, receives the precise answer, and then presents it to the user. Thus, intelligent agents not only benefit from the inherent abilities of language models but also increase their accuracy and efficacy in performing various tasks by using specialized tools.

**1.1 Problem Statement and the Goals of this Research**

These advancements motivated us to build a framework that can serve as an Educational Assistant for the Robotics course, so that students worldwide will be able to interact with it easily and inquire about their robotics problems (with a focus on robotic arms). Such a framework can even be utilized as a Teacher Assistant in schools and universities and interact with the user just like a language model. The significance of building this system lies in providing accessible education for everyone globally. Therefore, the objective of this research is to develop a highly modern educational assistance framework in the domain of robotic arms, based on intelligent agents driven by LLMs. In this study, the designed framework called "RoboSolver", by utilizing the Natural Language Processing (NLP), decision-making, thinking, and advanced reasoning capabilities of LLMs (while being equipped with its own smart, specialized tools), helps the system to interpret high-level (and even colloquial) user instructions regarding manipulators and generate appropriate responses in real-time. Therefore, the proposed framework, by leveraging its comprehensive suite of dedicated robotics tools and the advanced capabilities of LLMs, is capable of intelligently solving an acceptable range of problems posed in classic robotics courses. These integrated tools encompass the most useful concepts in Robotics courses:

- Parameter Extraction: Tools for derive necessary parameters for robot modeling from descriptive text or even images of the robotic arm

- Kinematic Analysis: Tools for solving forward kinematics analytically and numerically, as well as inverse kinematics numerically

- Motion Analysis: Tools for computing the Jacobian matrix, and the velocity and acceleration of key points on the robot

- Simulation: Tools for 3D robot simulation within environments such as Swift or Matplotlib
- Control: Tools for velocity control of the joints or the end-effector, as well as trajectory control for guiding the robot from a specified point toward a desired target

Consequently, the intended framework is able to take robotic arm problems (either in text or image format), extract the necessary robot modeling parameters from them; and, based on the user's query, perform the calculation of forward/inverse kinematics, the Jacobian, velocity and acceleration analysis, as well as the simulation and motion control of the robot, all in a fully autonomous manner. To the best of our knowledge, this is the first framework to possess such a comprehensive set of features. Figure 1 illustrates an example of a query-and-answer interaction between the user and the RoboSolver framework. The details of this framework will be described in the Method section.

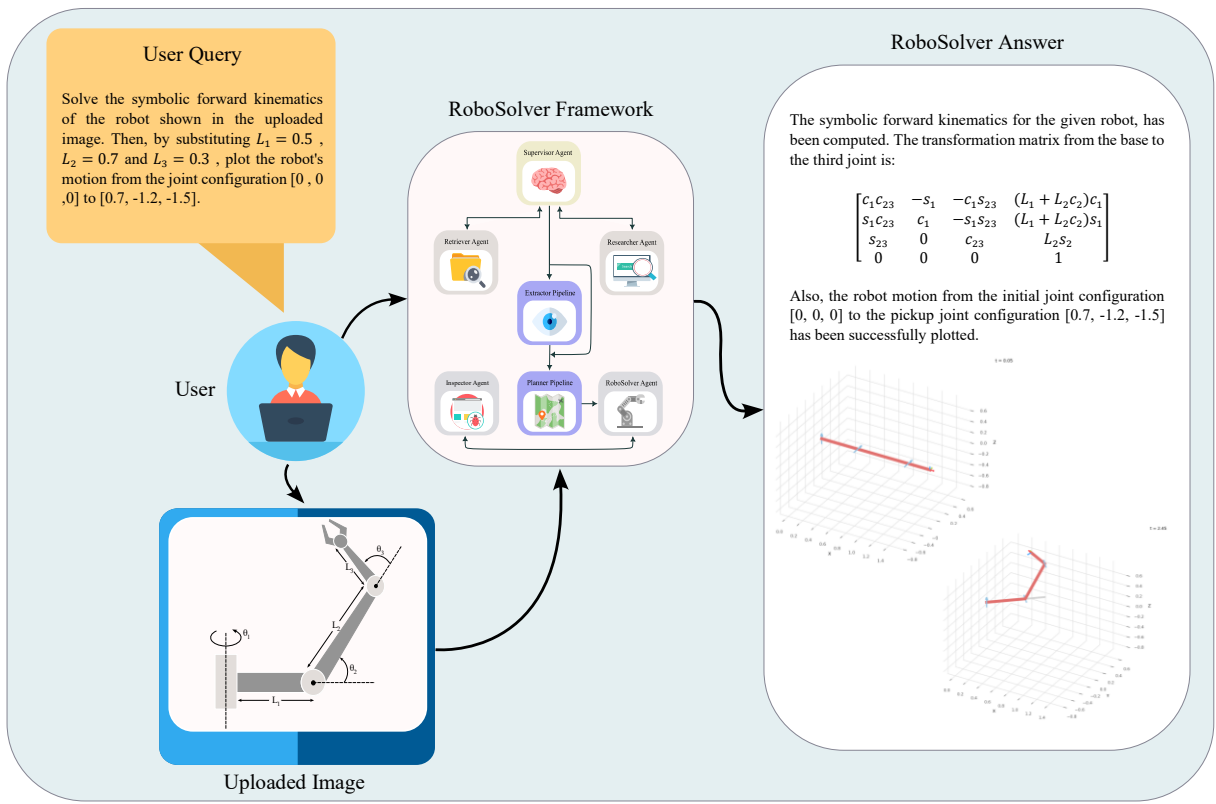

Figure 1: an example of a query-and-answer interaction between the user and the RoboSolver framework.

# 2 RELATED WORKS

## 2.1 Introduction

In the book "*Artificial Intelligence: A Modern Approach"*, an agent is defined as anything that can perceive its environment through sensors and act upon that environment through effectors [9]. Furthermore, Wikipedia states that in the context of Artificial Intelligence, an intelligent agent is an entity that independently perceives its surrounding environment, takes appropriate actions to achieve its goals, and has the ability to improve its performance through machine learning or knowledge acquisition [10]. The key concept is that agents receive feedbacks from the environments. The environment with which an agent can interact may be physical (such as the surroundings of a robot or an autonomous vehicle), digital (like interacting with operating systems and software, such as Siri), or even human (such as the interaction of chatbots with human users). Among the various types of environments with which agents can interact, the human environment holds a special place; chatbots are among the most prominent examples of this type of interaction. The first known chatbot, ELIZA, was developed in 1966 by Joseph Weizenbaum at the MIT Artificial Intelligence Laboratory [11]. This system operated based on a set of simple, predefined rules. Its architecture relied on rule-based patterns (such as *if-then* logic) and was only capable of responding within highly specific and constrained domains. These types of chatbots required manual design and programming and were ineffective in handling complex or out-of-scope scenarios. Subsequent technological advancements—particularly in the field of recurrent neural networks—led to a shift away from rule-based frameworks toward data-driven learning approaches. This transition enabled the development of systems capable of automatically learning patterns from large-scale textual datasets, allowing them to perform effectively across a broader range of domains without direct human intervention. For instance, in 2016, Jiwei Li and colleagues developed an experimental conversational model based on LSTM networks [12]. The model was trained on 23 million short Twitter conversations, enabling it to learn common patterns in everyday dialogues and generate contextually relevant responses. However, due to the nature of the training data, the model became biased toward the brief style of Twitter conversations, often producing short responses during interactions. On the other hand, incorporating longer conversations into the training process to mitigate this bias was not feasible, as LSTM networks—despite significant advancements—still faced inherent limitations in handling long-term dependencies.

Finally, with the advent of the Transformer architecture, new language models entered the arena that were capable of effectively retaining long-term information. This provided a smoother, more accurate, and more natural experience in user interaction. On the other hand, since these language models can communicate with their environment (which, in this case, is human), perceive it, and perform actions based on it—such as generating responses—they can be considered a type of agent based on the classic definition provided. However, the convention in the literature is that a language model is only recognized as a true intelligent agent when it leverages tools for active interaction with the environment [13] (such as web search, file reading, or code execution). Furthermore, receiving a "role-based prompt," which specifies the agent's role, objective, and scope of action, is essential for correctly directing the agent's decision-making and action.

Given the nascent nature of this research domain, a review of the background can provide a better understanding of the available opportunities. Subsequently, the literature related to agents and research conducted in the field of education will be examined. This literature review will analyze and evaluate the strengths, weaknesses, and existing research gaps in previous studies to clearly delineate the position and necessity of the current research. The research conducted in this study can be categorized into two main axes. The first section examines studies that address the fundamental question of whether LLMs can play an effective role in education, particularly in Mechanical Engineering education, which necessitates a deep understanding of technical and analytical concepts. The second section reviews studies related to the application of agents and language models in solving engineering problems and education within this domain. Finally, relying on a comprehensive analysis of these two research categories, the objective, position, and trajectory of the present study will be determined.

## 2.2 Background

In recent years, with the advances made in the field of artificial intelligence and the emergence of fundamental changes in various aspects of life, education has also begun to embrace more modern approaches. Moreover, it is anticipated that over the next decade, educational methods will undergo profound transformations, moving toward greater interactivity, the delivery of more effective visual content, and faster, easier, and more widespread accessibility. The realization of this vision will largely be enabled through the use of artificial intelligence, particularly advanced language models. Initially, the primary objective of this proposed research was the design and development of an educational assistant capable of solving problems related to robotic manipulators in a precise and generalizable manner. To this end, various studies and evaluations were conducted to identify an appropriate pathway toward achieving this goal. Ultimately, the results of these investigations indicated that one of the most suitable approaches for addressing robotic problems is the simultaneous use of language models and vision–language models (VLMs). This combination can, on the one hand, provide the ability to understand natural language and, on the other hand, offer high flexibility in solving various problems through the interpretation of visual inputs. However, prior to designing an educational assistant based on language models in the field of robotic arm, the most critical question is whether language models are capable of understanding a mechanical problem. As we know, in mechanical engineering education, the mere ability to comprehend textual information is not sufficient (however, it should be considered that these models are continuously progressing in different aspects such as education!). Thus, although language models demonstrate strong capabilities in understanding natural language, the requirements of specialized educational domains are far more complex and extensive than understanding simple text [14]. In other words, because mechanics involves foundational, spatial, and abstract concepts of its own, understanding it requires, in addition to comprehension skills, other abilities such as spatial visualization, mathematical computation, and reasoning [15]. In this regard, Bingling and colleagues conducted a study evaluating the inferential capabilities of ChatGPT-4 in the field of mechanical engineering. The results showed that the model may generate incorrect answers when solving certain mechanical problems, such as deriving equations, and by insisting on the correctness of these answers, it can potentially cause confusion for students, particularly beginners. These findings indicate that despite the remarkable textual processing capabilities of ChatGPT-4, its use in specialized engineering problems requires caution [14]. Therefore, measures must be taken to reduce such errors. For example, in operations involving symbolic equations, instead of relying on the inherent capabilities of a language model, one can use a library such as SymPy as a dedicated tool—a powerful Python library specifically designed for symbolic computation. The rationale behind this choice is that language models are fundamentally trained on next-token prediction within a textual sequence and, by design, lack the intrinsic computational structure required for precise and fully symbolic mathematical problem solving. In contrast, libraries such as SymPy, by employing rigorous symbolic and algebraic methods, are capable of performing complex operations such as simplification, differentiation, and equation solving with complete accuracy. In a more in-depth and analytical study, Tian and colleagues systematically evaluated the capabilities of ChatGPT 3 and 4 in solving mechanical engineering problems. To this end, a comprehensive test consisting of 126 questions—designed by a group of experts in the field—was employed to assess the models' level of proficiency. In a thorough evaluation, the performance of these two models in solving mechanical engineering problems was compared with that of domain experts, including faculty members from mechanical engineering departments. The results of this study indicated that the aforementioned models outperformed university professors in many specialized subfields [16]. It is worth noting, however, that the test in question was administered in a closed-book format—conditions that naturally favor large language models. These models, by leveraging their extensive pretrained knowledge, possess the ability to accurately retrieve memorized information such as definitions, laws, and even less common formulas. In contrast, a professor—despite having the ability to understand and solve problems—may not recall certain formulas or memorized details due to the passage of time. Therefore, in situations where access to external resources is restricted, the superior accuracy of language models compared to humans who rely on memory is natural and expected. Moreover, the results indicated that ChatGPT models, when prompted to first provide their step-by-step reasoning and logical explanation (chain of thought [17]) and then present the final answer based on that reasoning, demonstrated a higher level of accuracy and logical reasoning in the subjects of dynamics and statics. In addition, when the prompt for ChatGPT-4 explicitly specified the domain in which the model was expected to answer questions, higher accuracy was achieved in the evaluations. This is because the model's distribution shifts toward the appropriate region of the problem space, enabling it to respond with greater expertise in the specified domain. Overall, the findings of this study—which indicate that ChatGPT-4 demonstrates higher accuracy in answering questions compared to university faculty—highlight the remarkable potential of this model to serve as an intelligent, precise, and reliable assistant in the field of engineering. Continuing the examination of language models' capabilities in solving mechanical engineering problems, the MechAgents framework offers an innovative example of leveraging multi-agent language-model collaboration for the automated solution of mechanical problems, particularly in the domain of elasticity [8]. In this framework, a group of language models functions as agents with specialized roles—such as planner, coder, executor, and critic—working interactively to write and execute finite element method (FEM) code, analyze the resulting outputs, and correct any errors that arise. This architecture, implemented using GPT-4 and the AutoGen platform, has been able to correctly solve a wide range of problems in solid mechanics without human intervention, demonstrating that task decomposition, reciprocal criticism, and dynamic interaction among agents lead to increased accuracy and coherence. One of the key features of this architecture is the role of the critic agent in the process of validating the responses. This agent evaluates the consistency of numerical or graphical outputs from simulations with physical principles, and even when the code runs without error, it can detect hidden anomalies (such as inconsistencies in the stress pattern) and provide appropriate feedback to other agents for correction. This level of self-monitoring goes beyond the capabilities of individual agents and is one of the major advantages of a multi-agent structure. Nevertheless, this framework is not without limitations. First, although the AutoGen platform supports the use of external functions (such as invoking specialized tools), this capability was not utilized in the study, and the approach relied solely on code generation and execution by the language model itself—an aspect that may become inefficient when dealing with more complex problems. Second, in the architecture of the AutoGen platform, all agents read each other's messages. While this leads to comprehensive oversight of the system's performance by all agents, it may be suboptimal in larger tasks or at broader scales in terms of token consumption. In another study conducted by Zahedifar et. al., an LLM-Agents-Controller was developed that is capable of solving classical control problems [18]. By employing multiple specialized agents, they intelligently distributed the tasks to enable their framework to operate effectively. For example, in this system, the user's query is first received by the Supervisor Agent, which determines—based on the type of question—which auxiliary agent should be invoked. If external data are required, the Retriever Agent extracts the relevant information, the Researcher Agent searches the internet for the necessary content, and the Reasoner Agent performs logical reasoning using methods such as chain-of-thought or tree-of-thought [19]. If no external assistance is needed, the Supervisor

Agent delegates the question to the Planner Agent, which determines the sequence of tools that should be executed. The Controller Agent then generates the response using the designated tools, and in the event of an error, the Debugger Agent resolves the issue. Subsequently, the generated answer is evaluated by the Critic Agent, and any required improvements are applied. Finally, the Communicator Agent delivers the response to the user in an appropriate format and style. The implementation of this framework represents an important step toward realizing the vision of fully automated educational architectures based on intelligent agents—systems capable of solving mechanical engineering problems comprehensively, from understanding the problem to delivering the final solution. However, one criticism of this architecture lies in the large number of agents involved in the problem-solving process, which increases system complexity, prolongs processing time, and leads to high token consumption. Therefore, by redesigning and reorganizing the agent structure—such as removing nonessential components or merging overlapping agents—a more optimized framework can be achieved; one that, while maintaining reliable performance, is also more efficient in terms of resource utilization. These efforts are being pursued while the automatic solution of dynamic and physical problems using language models is still in its early stages. Given the emerging nature of LLM agents and the inherent complexity of automating the solution of robotic and dynamic problems, the existing work in this domain remains scarce and limited. For example, Du et al., in an innovative study, addressed the problem of discovering governing equations of physical phenomena directly from experimental data. In their approach, instead of relying on predefined physical laws or classical mathematical analysis, LLMs are employed to automatically propose equations that can describe the system's dynamics based on a set of numerical observations. This method replaces the traditional and time-consuming process of deriving equations with the inferential capabilities of language models, opening new horizons for data-driven analysis of physical phenomena [20]. In another study conducted by Gerstmayr and colleagues, the Exudyn Python library and large language models were employed for dynamic modeling. In this approach, the user first provides a textual prompt describing the dynamic problem, thereby supplying the necessary information to the language model. The model then uses the Exudyn library to generate Python code and implement the system's dynamic model [21]. Despite the novelty of this method, several limitations exist. First, preparing a precise and comprehensive prompt to fully describe the system can be a challenging and time-consuming task for the user. Second, their approach lacks the ability to directly interpret problems from images. Third, the study does not introduce any tool or function designed to optimize the interaction between the language model and Exudyn—an addition that could have improved overall efficiency and prevented many errors that typically arise during model-generated coding.

As shown, numerous studies in recent years have focused on the development of intelligent agents, particularly within the field of education. However, to the best of our knowledge, no framework based on LLM agents has yet been presented that is specifically capable of solving problems related to robotic arms and acting as an intelligent educational assistant for learners. Therefore, inspired by previous studies, we decided to design the RoboSolver framework, which is based on LLM agents and is capable of solving problems in the domain of manipulators. This framework will be capable of receiving textual and visual inputs (in contrast to the work of Gerstmayr et al., which relied solely on textual input) and using them for robot modeling. Once the robot model is obtained, it will also be able to perform analyses such as forward and inverse kinematics, calculation of key-point velocities and accelerations, as well as robot simulation and control.

## 3 METHODOLOGY

As discussed in the previous section, the remarkable advances in LLMs have significantly transformed research approaches in the development of autonomous educational agents. These agents can now, much like the conductor of an orchestra, coordinate and manage all stages of the problem-solving process in order to address the user's needs accurately and efficiently. In this regard, an appropriate foundation was established for designing a framework capable of serving as an intelligent educational assistant in the field of robotics. The proposed framework in this study will employ a multi-agent architecture, similar to the work of Zahedifar et al. [18], but with a key distinction: to optimize resource usage, the number of agents will be reduced. Moreover, unlike the MechAgents study [8], which utilizes the AutoGen platform, this framework will be implemented using LangChain, since in this platform only the necessary agents are activated at the appropriate time, eliminating the need for all agents to read one another's messages. Additionally, in contrast to the approach presented by Gerstmayr et al. [21], the agents in our framework will not be required to generate programming code from scratch. Instead, a set of predefined tools (developed by us) is made available to the agents, and each agent is responsible only for selecting the appropriate tool and determining the correct input needed to obtain the final answer. This approach not only substantially reduces the likelihood of errors in code generation and execution, but also allows the agents to focus on logical reasoning and goal-oriented decision-making rather than becoming entangled in technical implementation details—ultimately leading to a significant improvement in the model's accuracy. Consequently, the proposed framework can deliver faster, more optimized, and highly reliable performance compared to existing approaches.

On the other hand, to successfully handle a complex request, an agent must be able to decompose it into a set of smaller sub-tasks and then select and execute the appropriate tool for each step. The ReAct architecture was one of the first approaches to implement this process effectively and in a generalizable manner [22], thereby paving the way for subsequent research in this direction. In this method, the language model plays a central role in both the reasoning and action components. For each step of the problem, the model first analyzes the situation through linguistic reasoning and then selects and executes an appropriate action based on that reasoning. This process continues as an iterative loop until the agent reaches the user's final objective. One of the key features of this architecture is that if the model makes a mistake at any step or arrives at an incorrect result, it can detect the error by analyzing previous outcomes in the next iteration and subsequently adjust its decision-making path. For this reason, the agents in the proposed framework make decisions based on the ReAct architecture. The overall structure of our proposed framework is illustrated in the Figure 2. In the following sections, we introduce the components of this framework and examine its architecture in detail.

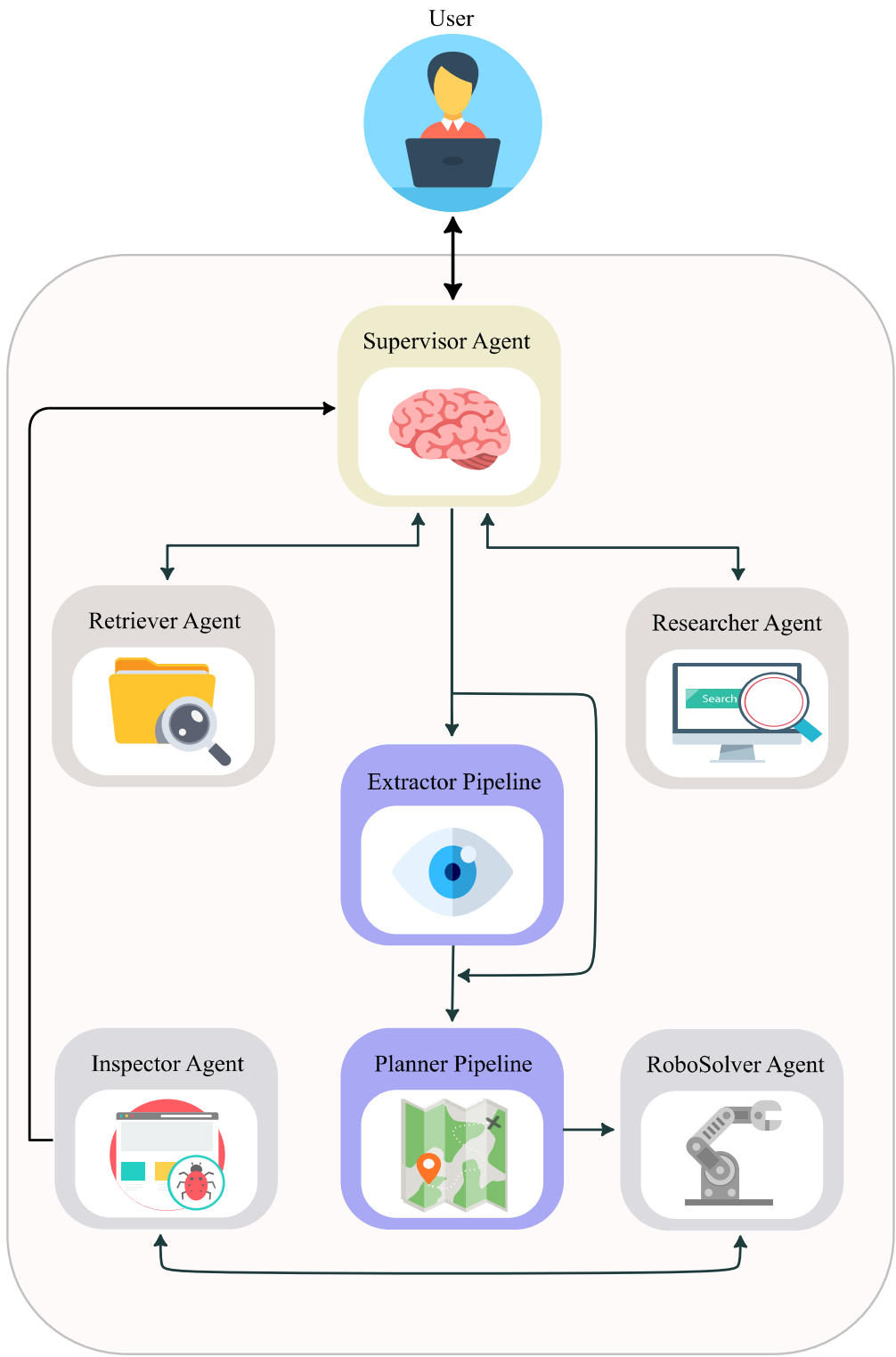

Figure 2: Architecture of the proposed RoboSolver framework.

## 3.1 Framework Architecture

**Supervisor Agent**

In this framework, the supervisor agent functions as the central brain of the system and also serves as the primary interface between the user and the other components of the system. Its secondary role is to establish and maintain an effective and dynamic linguistic interaction with the user so that it can accurately understand the user's request and ensure coherent communication between the human and the system. Once the user's request is clearly identified through this linguistic interaction, the agent's main task begins: it must select and activate the most appropriate agent or pipeline to address the user's request. Decision-making at this stage is based on the nature of the query, the type of input, and the objective of the problem, ensuring that the processing path is correctly determined and the user's needs are met accurately and efficiently. The supervisor agent has four decision-making routes at its disposal:

1- Researcher Agent

2- Retriever Agent

3- Extractor Pipeline

4- Planner Pipeline.

**Researcher Agent**

Whenever responding to the user's request requires searching reliable online sources, the supervisor agent invokes the researcher agent to perform this process using its dedicated tools. After gathering the necessary information and generating a response, the researcher agent returns the results to the supervisor agent, which then re-evaluates the situation—taking into account the newly obtained external information—to determine the most appropriate route for continuing the processing.

**Retriever Agent**

When the user provides a file (such as a PDF) and submits a request related to its content, the supervisor agent activates the retriever agent. This agent first extracts the relevant information from the provided source using its designated tools, and then the language model within the agent generates an accurate and context-appropriate response based on the retrieved data. Finally, the response is returned to the supervisor agent, which re-evaluates the situation—taking into account the newly obtained retrieved information—to decide on the next appropriate processing route.

**Extractor Pipeline**

This pipeline is activated by the supervisor agent when the user submits an image of a robotic arm or a textual description of its structure. Depending on the type of input, the pipeline uses a VLM for images and an LLM for textual descriptions, along with a set of predefined specialized prompts, to extract the parameters required for robot modeling. The traditional and widely used approach for modeling serial robots has been the Denavit–Hartenberg (DH) method. However, in 2007, Peter Corke introduced a new method called the Elementary Transform Sequence (ETS), which is simpler and more intuitive than the conventional approach [23]. Based on the qualitative evaluations conducted in this study, it was observed that even VLMs exhibit better compatibility with the ETS method and demonstrate superior performance in extracting its parameters from images of the robot compared to the Denavit–Hartenberg method. Thus, the prompts are designed so that the LLM or VLM is able to extract the ETS parameters from texts or images. Since possessing these parameters enables the complete modeling of robotic arms, this stage plays a pivotal role in the subsequent process of addressing the user's request, and any error introduced here can directly affect the accuracy of the final outcome. In the ETS method, the kinematic chain of the robot is modeled as a sequence of elementary transformations consisting of basic rotations and translations along or about the coordinate axes. The following section examines how this approach is employed in the kinematic modeling of a robotic arms.

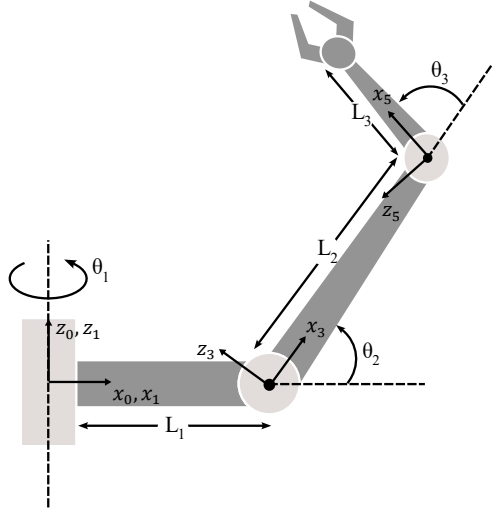

Figure 3: A view of a robotic arm.

Figure 3 illustrates a robotic arm. To obtain the Elementary Transform Sequence of this robot, the base (zero) coordinate frame is assumed to be fixed at the first joint. Initially, the first coordinate frame is placed exactly on top of the base frame. As shown in the figure, this frame undergoes a rotation of $\theta_1$ about the $z_0$ axis (in other words, the $x_1$ axis is fixed to link $L_1$ and rotates along with it). Next, a translation of $L_1$ along the $x_1$ axis is applied to reach the second coordinate frame located at the second joint. This is followed by a rotation of $-\theta_2$ about the $y_2$ axis and another translation of $L_2$ along the $x_3$ axis to reach the final joint. At the last joint, a rotation of $-\theta_3$ about the $y_4$ axis is performed so that the resulting coordinate frame aligns with the end-effector frame, whose $x$-axis coincides with the direction of link $L_3$. Thus, the resulting ETS can be written as:

$$ETS = R_z(\theta_1) \to T_x(L_1) \to R_y(-\theta_2) \to T_x(L_2) \to R_y(-\theta_3) \tag{1}$$

The VLM or LLM within the extractor pipeline is responsible for using the provided instructions to derive the ETS from images or textual descriptions. Subsequently, this information—together with the user's most recent request—is passed as inputs to the planner pipeline, so that the planning stage of the problem-solving process can begin.

**Planner Pipeline**

Based on the data received from the supervisor agent or the extractor pipeline, this pipeline determines which of the available tools in the Robosolver agent—consisting of both analytical and numerical components—are required to address the user's problem. It then optimally determines the order in which these tools should be executed. Subsequently, the output of the planner pipeline is delivered to the Robosolver agent as a decision-making guideline. This allows the Robosolver agent—without needing to perform any tool-selection analysis—to focus solely on determining the appropriate inputs and executing the required tools efficiently, ultimately resulting in more effective and accurate performance. An important point in this section is that only the tools selected by the planner agent are provided to the Robosolver agent. This ensures that the LLM within the Robosolver agent receives a more concise input, as descriptions of tools that are not intended to be used are omitted. Consequently, the number of input tokens is reduced, leading to lower API costs for the LLM. Moreover, with a smaller and more focused input, the model can concentrate more effectively on the problem at hand, ultimately improving its accuracy.

**RoboSolver Agent**

In this stage, the Robosolver agent, guided by the planner pipeline and based on the user's query, determines the input for the selected tool and invokes it. After observing the tool's output, the agent leverages its language-based reasoning capability to analyze the result and, accordingly, identify the appropriate input for the next tool and execute it. This process continues as an iterative cycle of reasoning and action until the Robosolver agent determines that the final answer has been successfully obtained.

In our platform, a total of 16 tools have been designed for this agent, categorized into two groups: analytical and numerical. The analytical tools were developed from scratch using the SymPy library, whereas the numerical tools were built upon the classes and functions provided by the Robotics Toolbox for Python library. As mentioned earlier, once the ETS parameters are available, the robot can be modeled with ease. Therefore, after these parameters are obtained from the extractor pipeline and passed to the Robosolver agent, the agent can readily utilize the available tools to perform the following operations:

1. Forward and inverse kinematics computation

2. Jacobian calculation

3. Velocity and acceleration computation

4. Robot 3D simulation

5. Robot control

For example, in the forward-kinematics tool, it is sufficient to use the extracted ETS, as it represents an ordered sequence of elementary transformations that directly enables the computation of forward kinematics. By following this sequence and consecutively multiplying the corresponding transformation matrices for each joint or link, the final transformation matrix of the robot is obtained. For instance, the forward kinematics of the robot shown in Figure 3, based on the ETS extracted in Eq. (1), is expressed in Eq. (2):

$$\begin{aligned}Forward\ Kinematics &= \begin{bmatrix} c_1 & -s_1 & 0 & 0 \\ s_1 & c_1 & 0 & 0 \\ 0 & 0 & 1 & 0 \\ 0 & 0 & 0 & 1 \end{bmatrix} \begin{bmatrix} 1 & 0 & 0 & L_1 \\ 0 & 1 & 0 & 0 \\ 0 & 0 & 1 & 0 \\ 0 & 0 & 0 & 1 \end{bmatrix} \begin{bmatrix} c_2 & 0 & -s_2 & 0 \\ 0 & 1 & 0 & 0 \\ s_2 & 0 & c_2 & 0 \\ 0 & 0 & 0 & 1 \end{bmatrix} \\ &\quad \begin{bmatrix} 1 & 0 & 0 & L_2 \\ 0 & 1 & 0 & 0 \\ 0 & 0 & 1 & 0 \\ 0 & 0 & 0 & 1 \end{bmatrix} \begin{bmatrix} c_3 & 0 & -s_3 & 0 \\ 0 & 1 & 0 & 0 \\ s_3 & 0 & c_3 & 0 \\ 0 & 0 & 0 & 1 \end{bmatrix} = \begin{bmatrix} c_1 c_{23} & -s_1 & -c_1 s_{23} & (L_1 + L_2 c_2) c_1 \\ s_1 c_{23} & c_1 & -s_1 s_{23} & (L_1 + L_2 c_2) s_1 \\ s_{23} & 0 & c_{23} & L_2 s_2 \\ 0 & 0 & 0 & 1 \end{bmatrix}\end{aligned} \quad (2)$$

To obtain the velocity and acceleration, it is sufficient to use the position components of the final transformation matrix and differentiate them once and twice with respect to time, respectively. The first derivative yields the velocity of the desired point, while the second derivative provides its acceleration. Furthermore, robot simulation is carried out using the Swift and Matplotlib environments, both of which are well supported by the Robotics Toolbox for Python. In the simulation section, specialized tools have been developed for the Robosolver, through which the robot's behavior can be observed by specifying either the joint velocities or the end-effector velocity. In the Swift environment, the robot's motion trajectory is controlled using a Resolved-Rate Motion Controller (RRMC). To support this, dedicated tools have been developed for the Robosolver agent, enabling it to guide the robot's end-effector from an initial point toward a static or moving target. This task is carried out entirely within the simulation environment and can be visually presented to the user. For numerical inverse kinematics, the solvers provided by the Robotics Toolbox for Python are employed. These solvers include various methods such as Newton–Raphson, Gauss–Newton, and several variants of the Levenberg–Marquardt algorithm, including chan, sugihara, and wampler. All of these methods operate based on the RRMC concept and are capable of solving the inverse kinematics problem within only a few tens of microseconds. For details on Jacobian computation, the implementation of RRMC, and the numerical inverse-kinematics procedure, refer to [24] and [25].

**Inspector Agent**

The final response generated by the Robosolver agent is passed to the inspector agent, whose role is to verify whether the user's request has been correctly fulfilled. In other words, if any errors occur during tool execution or if inconsistencies arise between the user's intended goal and the output produced by the Robosolver agent, the inspector agent provides corrective feedback to the Robosolver, prompting the problem-solving process to be repeated with the necessary improvements. Once the final response is validated in terms of accuracy and consistency, the inspector agent forwards the result to the supervisor agent, which then presents it to the user in an organized manner. Finally, the response—along with the user's original query—is stored in the internal memory of the supervisor agent. This memory helps the agent in future interactions, enabling it to correctly recognize any implicit or explicit references the user makes to previous questions or answers, and to effectively incorporate that information when interpreting the user's new request.

## 3.2 Framework Evaluation

To evaluate the developed framework, it is first necessary to define an appropriate set of assessment metrics. For this purpose, the evaluation criterion previously employed in the study by Zahedifar et al. [18] was adopted, but redesigned with modifications tailored to the requirements of the RoboSolver Framework. In the redesigned version, each agent and each pipeline is scored based on whether it successfully performs its assigned task: a score of 1 is assigned if the task is completed correctly, and a score of 0 otherwise. In the following section, these metrics are discussed in detail.

### 3.2.1 Metrics
#### 3.2.1.1 Supervisor Agent Score

This score evaluates the effectiveness of the supervisor agent in managing the problem-solving workflow. It reflects the agent's ability to understand the user's intent through dynamic interaction and, subsequently, to correctly route the process by selecting and arranging the appropriate agents and pipelines required to solve the problem (Eq. (3)).

$$\mathcal{M}_{Sup} = \frac{1}{N}\sum_{n=1}^{N} \frac{1}{2} [\mathbb{I}(User\ Interaction\ =\ Proper\ User\ Interaction) \\ + \mathbb{I}(Routed\ Next\ Agent/Pipline\ =\ Correct\ Next\ Agent/Pipline)] \quad (3)$$

In Equation (3), $N$ denotes the number of times the framework is queried. The indicator function, denoted by $\mathbb{I}(.)$, returns a value of 1 if the specified condition is satisfied and 0 otherwise. If the interaction with the user is carried out correctly, leading to an accurate interpretation of the user's request and consequently the correct selection of the next agent or pipeline, this score will be assigned a value of 1. However, if either of these two criteria is not properly satisfied, the score will be 0.

#### 3.2.1.2 Extractor Pipeline Score

This score evaluates the accuracy of this pipeline in extracting and providing the ETS parameters.

$$\mathcal{M}_E = \frac{1}{N}\sum_{n=1}^{N} \mathbb{I}\,[ETS\ Extracted\ =\ Correct\ ETS] \quad (4)$$

According to Equation (4), If the extracted parameter is correct, the indicator function $\mathbb{I}(.)$ returns 1; otherwise, it returns 0.

#### 3.2.1.3 Planner Pipeline Score

This score assesses whether the pipeline has performed appropriately in selecting and ordering the necessary tools required to solve the problem.

$$\mathcal{M}_P = \frac{1}{N}\sum_{1}^{N} \mathbb{I}[Planned\ Tools\ =\ Correct\ Tool] \tag{5}$$

According to Equation (5), if the tools selected by this pipeline are both correctly chosen and arranged in the proper order, the indicator function $\mathbb{I}(.)$ returns 1; otherwise, it returns 0.

### 3.2.1.4 RoboSolver Agent Score

This score evaluates the effectiveness of the Robosolver agent in producing the correct answer as the core component of the final response.

$$\mathcal{M}_R = \frac{1}{N}\sum_{1}^{N}\frac{1}{\tau}\sum_{1}^{\tau} \mathbb{I}[\ RoboSolver\ Agent\ Answer\ =\ Correct\ Answer] \tag{6}$$

In Equation (6), $\tau$ represents the number of times the Robosolver agent is executed during a single run of the framework. If the final response produced by the Robosolver agent is correct, the indicator function $\mathbb{I}(.)$ returns 1; otherwise, it returns 0.

### 3.2.1.5 Judgement Score

This score measures the accuracy of the inspector agent in validating the final response produced by the Robosolver agent.

$$\mathcal{M}_J = \frac{1}{N}\sum_{1}^{N}\frac{1}{\tau}\sum_{1}^{\tau} \mathbb{I}[(Correct\ Answer\ =\ Accepted\ Answer) \lor (Incorrect\ Answer\ =\ Unaccepted\ Answer)] \tag{7}$$

In Equation (7), $\lor$ represents the logical OR function. According to this equation, if the inspector agent correctly verifies a correct answer or correctly rejects an incorrect one, the indicator function $\mathbb{I}(.)$ returns 1; otherwise, it returns 0.

### 3.2.1.6 Self-Correcting Score

This score evaluates the quality of the corrective feedback provided by the inspector agent. In other words, when an error occurs in the Robosolver agent, it assesses whether the feedback offered is appropriate for accurately resolving that error.

$$\mathcal{M}_{SC} = \frac{1}{N}\sum_{1}^{N}\frac{1}{\tau}\sum_{1}^{\tau}\mathbb{I}[(Error\ Fixing)] \tag{8}$$

In Equation (8), If the feedback provided successfully resolves the error, the indicator function $\mathbb{I}(.)$ returns 1; otherwise, it returns 0.

### 3.2.1.7 Completion Score

This metric evaluates the correctness and completeness of the framework's final answer with respect to the user's query.

$$\mathcal{M}_C = \frac{1}{N}\sum_{1}^{N}\mathbb{I}[(Framework\ Final\ Answer\ =\ Correct\ Answer)] \tag{9}$$

In Equation (9), if the framework's response is correct, the indicator function $\mathbb{I}(.)$ returns 1; otherwise, it returns 0.

### 3.2.1.8 Total Score

This score represents the overall performance of the agents within the Robosolver framework and is obtained from the average of all the metrics discussed, except for $\mathcal{M}_C$.

$$\mathcal{M}_T = \frac{1}{6}(\mathcal{M}_{Sup} + \mathcal{M}_E + \mathcal{M}_P + \mathcal{M}_R + \mathcal{M}_J + \mathcal{M}_{SC}) \tag{10}$$

## 3.2.2 Evaluation Procedure

To evaluate the developed framework, three benchmark tests were designed. The first test consisted of textual descriptions of ten serial robots, while the second test included images of the same robots. These robots were selected from reference [26]. The textual descriptions were written by us and are presented in Figure 4. Moreover, due to the low quality of the images provided in [26], the corresponding images were redrawn and are shown in Figure 5.

The robot begins with Joint 1, which provides a rotation of $\theta_1$ about the global Z-axis, causing the entire structure to rotate from this joint, the mechanism rises a distance $L_1$ along the local Z-axis to reach Joint 2, which rotates by $-\theta_2$ about its local y-axis. From Joint 2, the structure extends forward a distance $L_2$ along the local x-axis to reach Joint 3, which likewise produces a rotation of $-\theta_3$ about its local y-axis. Finally, the robot advances another distance $L_3$ along the local x-axis, where the end-effector is positioned.

The robot begins with Joint 1, which provides a rotation of $\theta_1$ about the global Z-axis, rotating the entire assembly. From this joint, a link of length $L_1$, aligned with the local x-axis, connects to Joint 2, which rotates by $-\theta_2$ about its local y-axis. Continuing along Link 2, of length $L_2$, also oriented with the local x-axis, the structure reaches Joint 3, which produces a rotation of $-\theta_3$ about its local y-axis, similar to the second joint. Finally, the mechanism extends a distance $L_3$ along the local x-axis, where the end-effector is positioned.

The mechanism begins with Joint 1, which provides a rotational motion of $-\theta_1$ about the global Y-axis. From this joint, the structure extends a distance $L_1$ upward along the local Z-axis to reach a prismatic joint. This joint provides a translational motion along the local X-axis; in other words, the length of the sliding segment is $d_2$, corresponding to its maximum travel distance along that axis. At the end of this sliding segment lies Joint 3, which is a revolute joint rotating by $+\theta_3$ about the local Y-axis. Finally, Link 3, with a length of $L_3$, extends along the local X-axis and carries the end-effector at its tip.

The mechanism starts with a first joint that performs a rotation of $\theta_1$ about the global Z-axis. From this joint, the structure extends $L_1$ upward along the local Z-axis to reach the second joint, which rotates by $-\theta_2$ about the local y-axis. Next, the robot advances $L_2$ along the local Z-axis to reach a prismatic joint. This prismatic joint provides a translational motion of $d_3$ along the local x-axis, through which the sliding segment of length $d_3$ carries the end-effector at its tip.

The robot begins with a first joint that performs a rotation of $\theta_1$ about the global Z-axis. From this joint, the structure extends a distance $L_1$ along the local Z-axis to reach the second joint, which rotates by $-\theta_2$ about the local y-axis. Next, the mechanism advances a distance $L_2$ along the local Z-axis to reach the third joint. This joint produces a rotation of $-\theta_3$ about its local y-axis, causing the third link—of length $L_3$ and oriented along the local Z-axis—to rotate accordingly. The end-effector is mounted at the tip of this final link.

The mechanism starts with a first joint that performs a rotation of $\theta_1$ about the global Z-axis. Mounted on this joint is a prismatic joint that provides a translational motion of $d_2$ along the local Z-axis. Following this, another prismatic joint is connected in series, allowing an additional translation of $d_3$ along the local x-axis.

The mechanism begins with a prismatic first joint that provides a translation of $d_1$ along the global Z-axis. Mounted above it is a revolute joint that rotates by $\theta_1$ about its local Z-axis. From this joint, the structure extends a link of length $L_2$ along the local x-axis to reach the next revolute joint. This second revolute joint rotates by $\theta_2$ about the local Z-axis and drives the final link, $L_3$, which is aligned with the local x-axis and carries the end-effector at its tip.

The mechanism begins with a prismatic joint that provides a translational motion of $d_1$ along the global Y-axis. Following this, a second prismatic joint enables a translation of $d_2$ along the local X-axis. On top of these, a third prismatic joint is mounted, which produces a translation of $d_3$ along the local Z-axis. Attached to the sliding part of this last joint is a fixed link of length $L_4$, aligned with the local X-axis, and the end-effector is positioned at its tip.

The mechanism starts with a first revolute joint that rotates Link 1, which has a length of $L_1$ and is aligned with the local x-axis, by an angle of $-\theta_1$ about the global y-axis. At the end of this link, the second revolute joint is located. It operates in the same manner as the first joint, rotating Link 2, of length $L_2$, by an angle of $-\theta_2$ about its local y-axis. Following this, a third revolute joint is connected at the end of Link 2, again rotating Link 3, of length $L_3$, by $-\theta_3$ about its local y-axis. The end-effector is mounted at the tip of Link 3.

The robot begins with a revolute joint that rotates by an angle $\theta_1$ about the global Z-axis, driving the rotation of the entire structure. From this joint, the mechanism extends a distance $L_1$ upward along the local Z-axis to reach a prismatic joint. This prismatic joint provides a translational motion of $d_2$ along the local X-axis, and The end-effector is mounted at the tip of the sliding segment corresponding to $d_2$.

Figure 4: Textual descriptions of the benchmark robots.

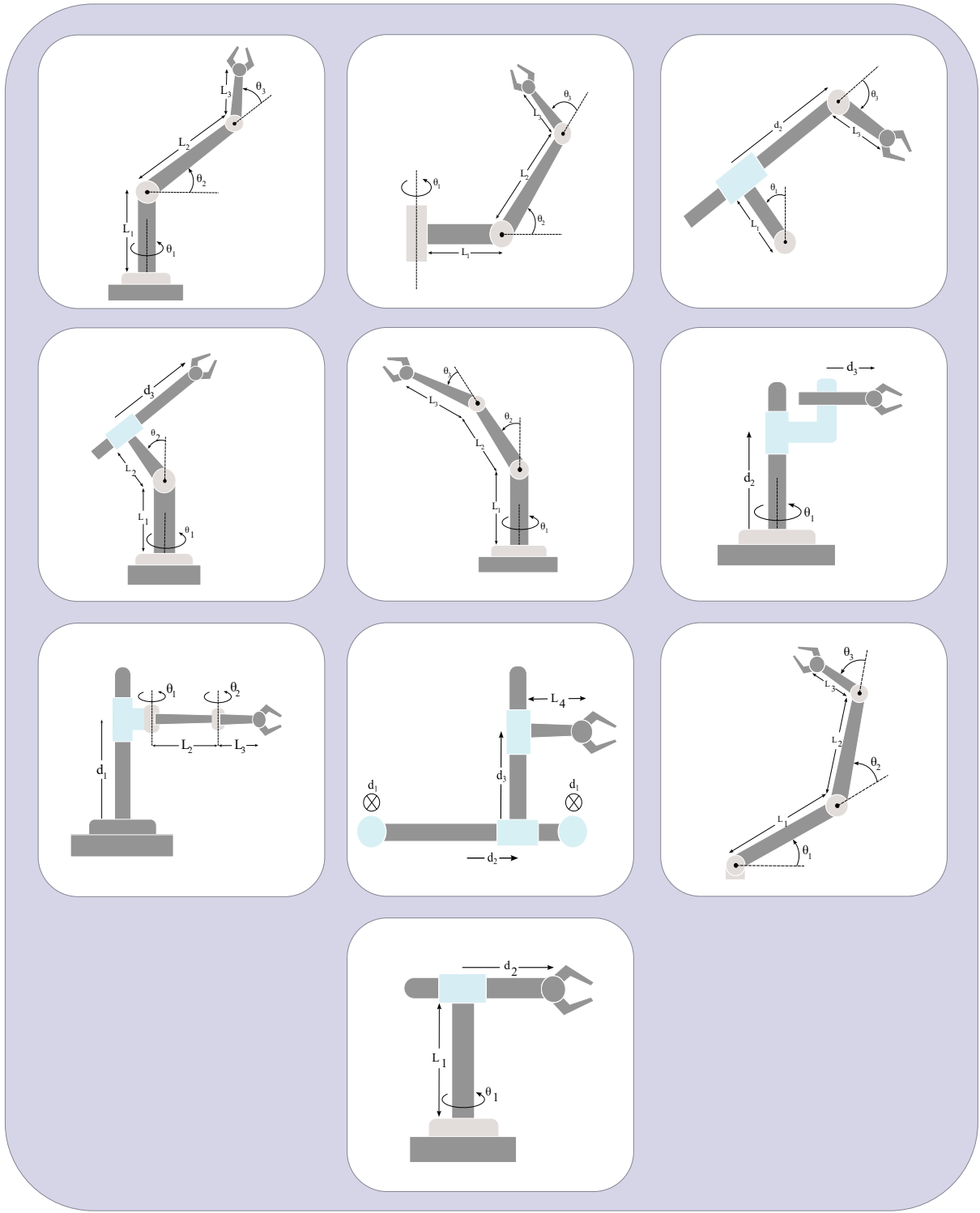

Figure 5: Visual representations of the benchmark robots.

In both tests, the framework was asked to compute the forward kinematics of each robot. This type of evaluation was chosen because the forward kinematics of a robotic arm effectively represents its fundamental model; Once it is obtained, performing other robotic tasks is easy. An important consideration in solving image-based problems by the framework is that, for a VLM to correctly interpret the robot image, the image must be captured in a two-dimensional, planar view (such as a side view), ensuring that all robot components lie within a single plane. Moreover, if the user's query is image-based, it is essential that the Joint Orientations in the World Frame be provided to the framework by the user to improve the accuracy of the image analysis and yield more precise results. Finally, a third benchmark test was designed to assess the overall performance of the framework in handling intertwined tasks such as forward and inverse kinematics, velocity and acceleration computation, Jacobian calculation, and robot control and simulation. In the following, we present the results obtained from these three benchmark tests and analyze them.

## 4 RESULTS AND DISCUSSION

### 4.1.1 Benchmark Test 1

In Benchmark Test 1, comparisons were conducted at two levels. First, the performance of the developed framework equipped with three LLMs—GPT-4o, DeepSeek-V3.2, and Claude-Sonnet-4.5—was examined to determine the impact of the LLM type on the framework's response quality. Second, the performance of these enhanced frameworks was compared with the raw versions of the same LLMs, that is, without any integrated tools. In this way, both the effect of tool augmentation and the relative effectiveness of different LLMs within the framework were assessed. The outcomes of these comparisons in benchmark test 1 are presented in Tables 1 and 2, as well as Figures 6 to 8. It is worth noting that in this test, each of the questions was executed three times per version to improve the validity and reliability of the results. Given that six versions of the system were evaluated, this resulted in a total of 180 runs.

Table 1: Comparison of Frameworks and Raw LLMs Performance Based on Evaluation Metrics in Benchmark Test 1.

| Model | Version | $\mathcal{M}_{Sup}$ | $\mathcal{M}_E$ | $\mathcal{M}_P$ | $\mathcal{M}_R$ | $\mathcal{M}_J$ | $\mathcal{M}_{SC}$ | $\mathcal{M}_C$ | $\mathcal{M}_T$ |
|---|---|---|---|---|---|---|---|---|---|
| GPT-4o | Framework | 1.00 | 1.00 | 1.00 | 1.00 | 0.97 | - | 0.97 | 0.99 |
| GPT-4o | Raw | - | - | - | - | - | - | 0.30 | - |
| DeepSeek-V3.2 | Framework | 0.93 | 1.00 | 0.93 | 0.93 | 0.93 | - | 0.93 | 0.94 |
| DeepSeek-V3.2 | Raw | - | - | - | - | - | - | 0.80 | - |
| Claude-Sonnet-4.5 | Framework | 1.00 | 0.77 | 0.77 | 0.77 | 1.00 | - | 0.77 | 0.86 |
| Claude-Sonnet-4.5 | Raw | - | - | - | - | - | - | 0.40 | - |

Table 2: Comparison of Average Token Usage, Execution Time, and Cost per Run for Frameworks and Raw LLMs in Benchmark Test 1

| Model | Version | Prompt Tokens | Completion Tokens | Total Tokens | Runtime (s) | Cost ($) |
|---|---|---|---|---|---|---|
| GPT-4o | Framework | 12277 | 907 | 13184 | 26.12 | 0.031 |
| GPT-4o | Raw | 168 | 1074 | 1242 | 15.82 | 0.011 |
| DeepSeek-V3.2 | Framework | 14307 | 1516 | 15823 | 49.34 | 0.004 |
| DeepSeek-V3.2 | Raw | 140 | 2662 | 2802 | 100.26 | 0.001 |
| Claude-Sonnet-4.5 | Framework | 20358 | 1623 | 21981 | 42.44 | 0.085 |
| Claude-Sonnet-4.5 | Raw | 363 | 1127 | 1490 | 14.14 | 0.018 |

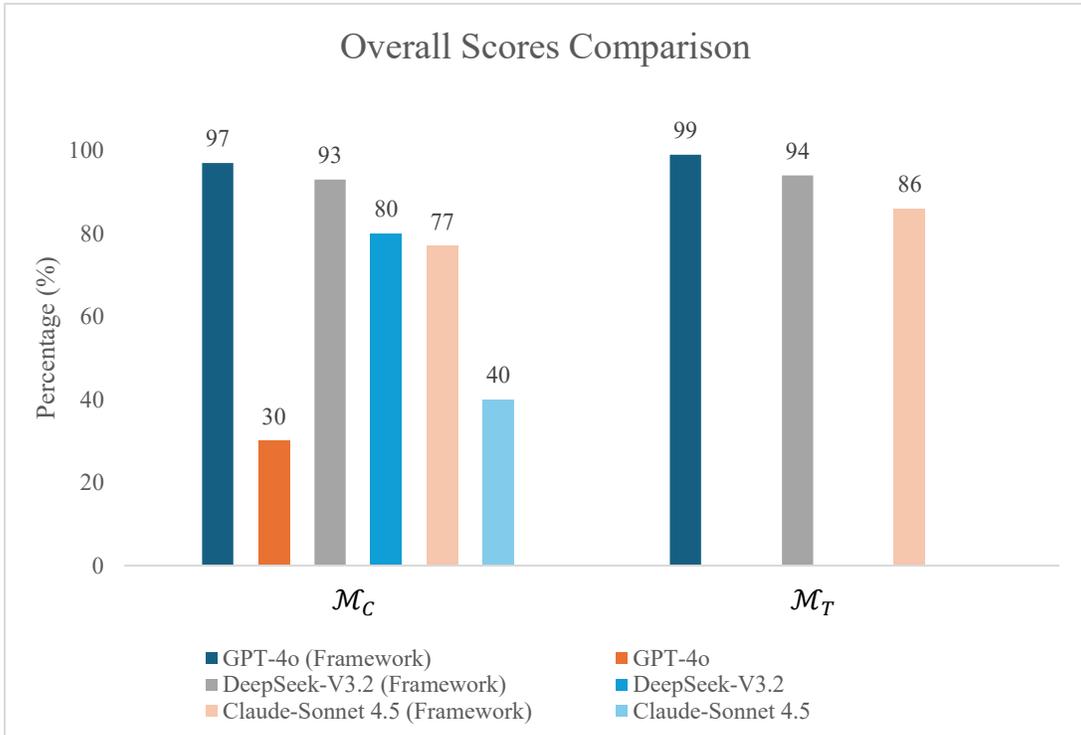

Figure 6: Comparison of Frameworks and Raw LLMs Overall Scores in Benchmark Test 1.

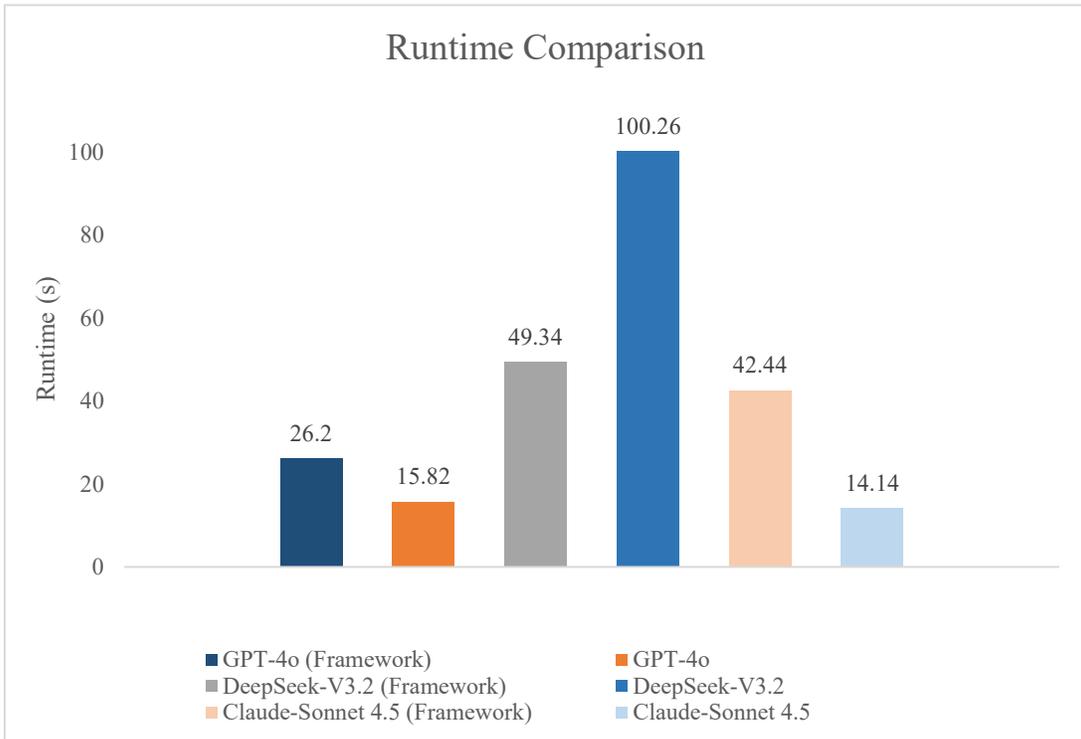

Figure 7: Comparison of Frameworks and Raw LLMs' Runtime in Benchmark Test 1.

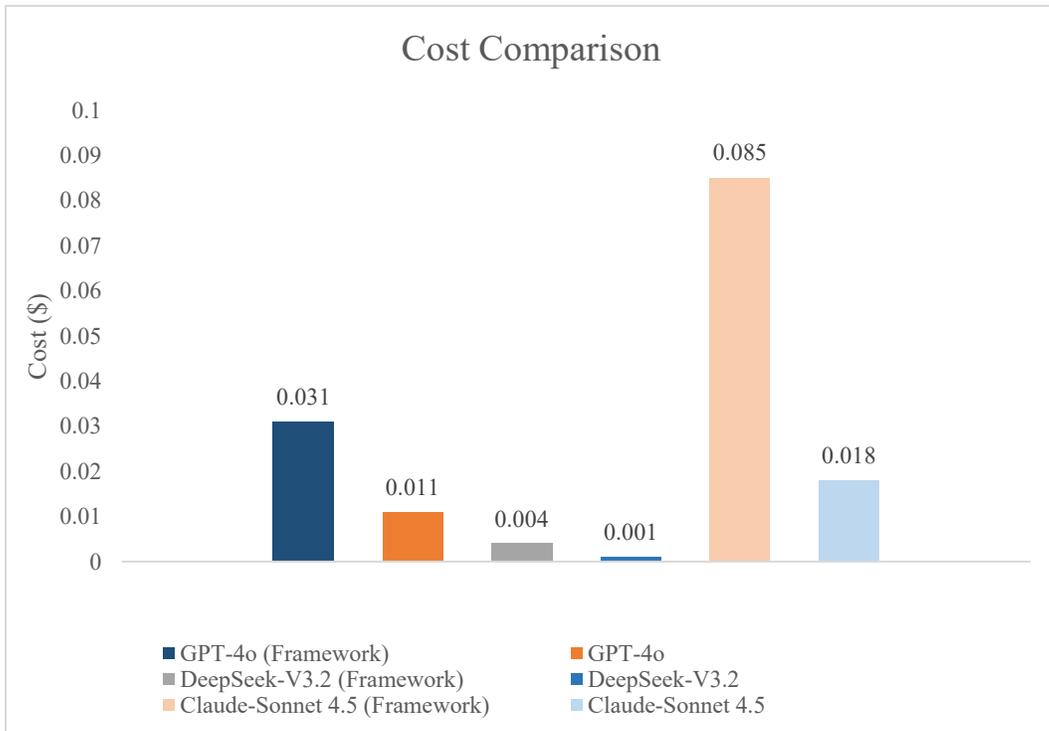

Figure 8: Comparison of Frameworks and Raw LLMs' Cost per Run in Benchmark Test 1.

In Table 2, Prompt Tokens and Completion Tokens represent the average number of input and output tokens processed by the model, respectively.

As shown, the $\mathcal{M}_C$ score achieved by our framework using the GPT-4o model reached 97%, representing a 67% improvement over the same model in its raw version. This score is also higher than those of the other frameworks built on DeepSeek-V3.2 and Claude-Sonnet-4.5. Notably, however, the framework employing DeepSeek-V3.2 achieved a score of 93%, while incurring approximately 87% lower cost compared to the GPT-4o-based framework. This suggests that despite slightly lower accuracy, DeepSeek-V3.2 may offer a highly cost-effective solution for commercial applications. Nevertheless, it is important to note that the response time of the framework using DeepSeek-V3.2 was nearly twice that of the GPT-4o-based framework, which could impact user experience in certain real-time applications. Another noteworthy observation is that the raw version of DeepSeek-V3.2 achieved an $\mathcal{M}_C$ score of 80%—only 13% lower than its tool-enhanced counterpart. Qualitative analysis shows that this model, even without tool support, was able to carry out complex matrix multiplications correctly by leveraging step-by-step reasoning. However, its response time was nearly twice that of the framework version using DeepSeek-V3.2, since in the absence of computational tools, the model had to perform matrix operations purely through linguistic inference. On the other hand, the framework based on Claude-Sonnet-4.5 exhibited lower accuracy despite having a cost nearly 20 times higher than that of the DeepSeek-V3.2-based framework. One of the primary reasons for this accuracy drop was that, in many cases where the framework failed to generate a final answer, the issue stemmed from the Supervisor Agent not passing the required input to other agents or pipelines. This lack of information exchange between agents often led to incomplete task execution and, ultimately, failure to arrive at a final answer. Given its lower cost and higher accuracy, the framework that uses DeepSeek-V3.2 is undoubtedly more favorable than the one based on Claude-Sonnet-4.5—especially considering that it is approximately 5 seconds slower in response time, a difference that is not significant enough to outweigh its overall advantage. Overall, the frameworks consistently outperformed the raw versions of the same LLM in terms of accuracy, demonstrating the effectiveness of the Robosolver framework in solving forward kinematics for robots described through textual input.

### 4.1.2 Benchmark Test 2

Since the framework that achieved the highest accuracy in Benchmark Test 1 was based on GPT-4o, in Benchmark Test 2, the framework exclusively employs the GPT-4o model alongside a VLM to ensure its performance in solving forward kinematics for visual problems is assessed at its maximum accuracy potential. Therefore, in this test, the Extractor pipeline employs the Gemini 2.5 Pro Preview VLM for image processing, while all other agents and pipeline continue to operate based on the GPT-4o model. The results of the comparison between the framework and the raw VLM in Benchmark Test 2, evaluated across 10 visual problem instances, are presented in Tables 3 and 4. It is worth noting that in this test as well, each question was executed three times, resulting in a total of 60 runs—30 for the framework and 30 for the raw model.

Table 3: Comparison of the Framework and Raw VLM Performance Based on Evaluation Metrics in Benchmark Test 2.

| Model | Version | $\mathcal{M}_{Sup}$ | $\mathcal{M}_E$ | $\mathcal{M}_P$ | $\mathcal{M}_R$ | $\mathcal{M}_J$ | $\mathcal{M}_{SC}$ | $\mathcal{M}_C$ | $\mathcal{M}_T$ |
|---|---|---|---|---|---|---|---|---|---|
| GPT-4o & Gemini 2.5 Pro Preview | Framework | 1.00 | 0.93 | 1.00 | 0.93 | 1.00 | - | 0.93 | 0.97 |
| Gemini 2.5 Pro Preview | Raw | - | - | - | - | - | - | 0.73 | - |

Table 4: Comparison of Average Token Usage, Execution Time, and Cost per Run for the Framework and Raw VLM in Benchmark Test 2.

| Model | Version | Prompt Tokens | Completion Tokens | Total Tokens | Runtime (s) | Cost ($) |
|---|---|---|---|---|---|---|
| GPT-4o & Gemini 2.5 Pro Preview | Framework | 16855.2 | 5486.2 | 22341.4 | 71.23 | 0.09 |
| Gemini 2.5 Pro Preview | Raw | 2897.83 | 13196.9 | 16094.73 | 128 | 0.14 |

A particularly noteworthy finding in this test is that the Gemini 2.5 Pro Preview model exhibits inherently agent-like behavior even in its raw version. In other words, the raw model is capable of autonomously invoking its internal tools when needed, without requiring any external framework. Despite this built-in capability, our proposed framework achieved an $\mathcal{M}_C$ score of 0.93—representing a 20% improvement over the direct use of Gemini 2.5 Pro Preview in its raw form. This was accomplished even under conditions where the raw model had an average response time of approximately one minute and an average cost about 55% higher than that of our developed framework. The increased response time and higher cost of the raw Gemini 2.5 Pro Preview model were mainly due to its tendency to initiate code generation as part of the problem-solving process. Overall, the results of this test demonstrated that the proposed framework has a clear advantage in solving visual forward kinematics problems, outperforming the raw model in terms of accuracy, response time, and cost. It is worth noting that the selection of Gemini 2.5 Pro Preview was motivated by the fact that, among existing VLMs, it was the only one that demonstrated adequate capability for accurate and reliable interpretation of robotic arm images.

### 4.1.3 Benchmark Test 3

The results of the two previous evaluations demonstrated that the developed framework performs well in solving the forward kinematics of robots when the queries are presented in either textual or visual form. As noted earlier, once the forward kinematics of a robot are obtained, constructing the robot's basic model becomes straightforward, which in turn enables the framework to perform a wider range of subsequent tasks. Accordingly, benchmark test 3 was designed to examine the framework's performance in a more comprehensive manner. This benchmark includes tasks such as computing velocities, accelerations, Jacobians, forward/inverse kinematics, and finally robot simulation and control. The benchmark consisted of 10 queries, 3 of which—together with the answers generated by the RoboSolver framework—are presented in Tables 5 to 7. Five of the question–answer pairs are provided in the Appendix, and one of them is also illustrated in Figure 1. Reviewing them is strongly recommended, as they offer a clearer understanding of the framework's capabilities and performance.

Table 5: Benchmark Test 3 – Query 1 and Corresponding Answers.

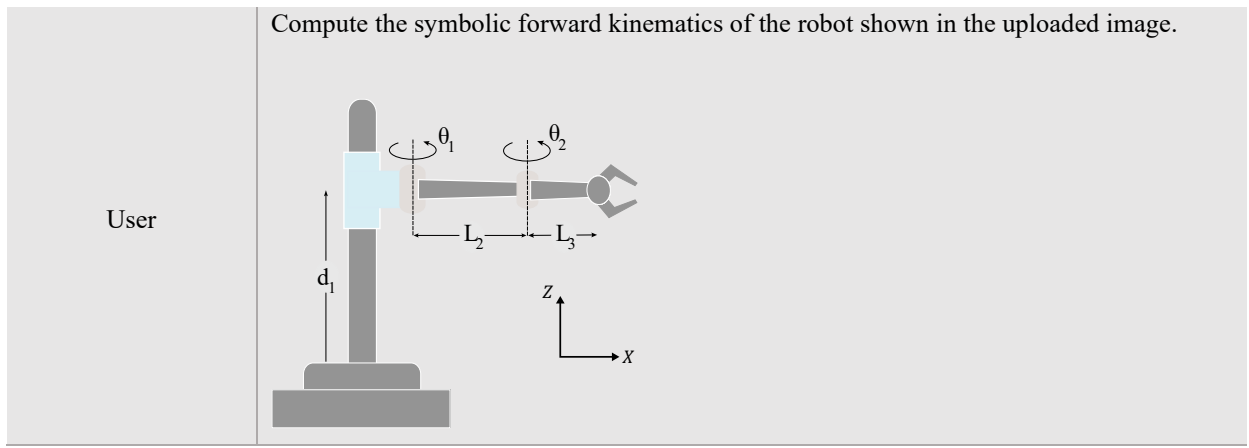

| User | Compute the symbolic forward kinematics of the robot shown in the uploaded image. |

| | | |
|---|---|---|
| | the joints' orientations of the robot are:<br>joint 1: along the Z-axis<br>joint 2: along the Z-axis<br>joint 3: along the Z-axis.<br><br>Then, by substituting $L_2 = 0.7$ and $L_3 = 0.5$, plot the robot motion from the joint configuration [0 , 0 ,0] to [0.7 , 0.5 ,0.7]. | |
| Supervisor | <Route to Extractor> | |
| Extractor | Based on the analysis of the robot manipulator, we can determine the elementary transforms (ET) for each joint and link in the kinematic chain. Here is the sequence of ETs:<br><br>1. **Joint 1 (Prismatic Joint):**<br>   - Translation along the Z-axis by a variable amount $d_1$.<br>   - ET: "tz($d_1$)"<br>2. **Joint 2 (Revolute Joint):**<br>   - Rotation about the Z-axis by an angle $\theta_1$.<br>   - ET: "Rz($\theta_1$)"<br>3. **Link between Joint 2 and Joint 3:**<br>   - Fixed translation along the X-axis by a length $L_2$.<br>   - ET: "tx($L_2$)"<br>4. **Joint 3 (Revolute Joint):**<br>   - Rotation about the Z-axis by an angle $\theta_2$.<br>   - ET: "Rz($\theta_2$)"<br>5. **Link from Joint 3 to End-Effector:**<br>   - Fixed translation along the X-axis by a length $L_3$.<br>   - ET: "tx($L_3$)" | |
| Planner | The user wants to perform symbolic forward kinematics on a robot defined by a sequence of elementary transforms (ETs). After obtaining the symbolic forward kinematics, the user wants to substitute specific values for the link lengths ($L_2 = 0.7$ and $L_3 = 0.5$) and then plot the robot's motion from an initial joint configuration to a final one.<br>To achieve this, the following steps are needed:<br>**Symbolic Forward Kinematics**:<br>Use the Symbolic_Forward_Kinematic_ET tool to compute the symbolic forward kinematics based on the provided ETs. | |

| | |
|---|---|
| | **Create Custom User-Defined Robot**:<br>Convert the symbolic ETS into a numerical model by substituting the given link lengths and use the create_custom_userdefined_robot tool to create a numerical model of the robot.<br>**Plot Custom Robot Motion**:<br>Use the plot_custum_robot_motion tool to plot the robot's motion from the initial joint configuration [0, 0, 0] to the final configuration [0.7, 0.5, 0.7]. |
| RoboSolver | The symbolic forward kinematics for the robot, defined by the sequence of elementary transforms, has been computed. The transformation matrix is as follows:<br><br>$$\begin{bmatrix} \cos(\theta_1 + \theta_2) & -\sin(\theta_1 + \theta_2) & 0 & L_2\cos(\theta_1) + L_3\cos(\theta_1 + \theta_2) \\ \sin(\theta_1 + \theta_2) & \cos(\theta_1 + \theta_2) & 0 & L_2\sin(\theta_1) + L_3\sin(\theta_1 + \theta_2) \\ 0 & 0 & 1 & d_1 \\ 0 & 0 & 0 & 1 \end{bmatrix}$$<br><br>After substituting the link lengths $L_2 = 0.7$ and $L_3 = 0.5$, the robot was numerically modeled and its motion was plotted from the initial joint configuration [0, 0, 0] to the final configuration [0.7, 0.5, 0.7]. The operation was performed on the robot with ID 'd03083e4-4bec-4d87-a431-af232d99decb'.<br><br>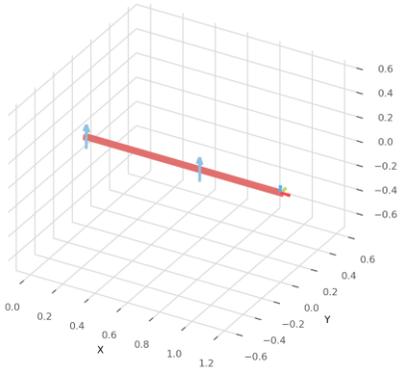 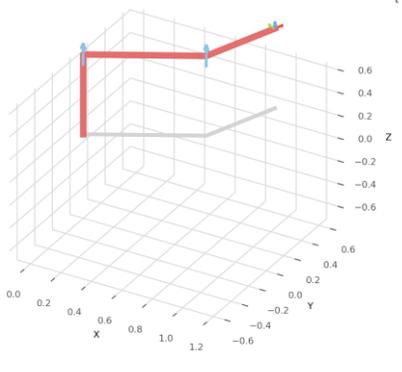<br>First Frame        Last Frame |

Table 6: Benchmark Test 3 – Query 2 and Corresponding Answers.

| | |
|---|---|
| User | Create an LBR robot and initialize it in the ready joint configuration.<br>Then, using the following desired end-effector positions in the world frame: |
| | $$\begin{bmatrix} 0.77 & -0.47 & 0.42 & 0.6 \\ 0.42 & 0.87 & 0.22 & -0.4 \\ -0.47 & 0 & 0.87 & 0.9 \\ 0 & 0 & 0 & 1 \end{bmatrix}$$ Using a position controller with a proportional gain of [3,3,3,3,3,3], simulate the robot from the ready joint configuration to the desired end-effector target. Also apply a velocity-control profile for motion execution. |
| Supervisor | <Route to Planner> |
| Planner | The user wants to simulate the position-based servoing of an LBR robot using a specific end-effector transformation matrix in the world frame. The task involves creating the LBR robot, initializing it in the ready joint configuration, and then simulating its motion using position-based servoing with a given proportional gain and velocity control profile. To achieve this, the following tools are needed: **create_robotictoolbox_robot**: This tool is used to create the LBR robot model from the predefined models available in the Robotics Toolbox library. **simulate_robot_motion_Position_based_Servoing**: This tool is used to simulate the robot's position-based servoing using the specified end-effector transformation matrix, initial joint configuration, proportional gain, and velocity control profile. |
| RoboSolver | The LBR robot was successfully created and initialized in the ready joint configuration. The position-based servoing was simulated using the specified end-effector transformation matrix in the world frame, with a proportional gain of [3, 3, 3, 3, 3, 3] and a velocity control profile. The operation was performed on the robot with ID 81084af4-40b3-4886-9b4e-0000ad0f9f29. 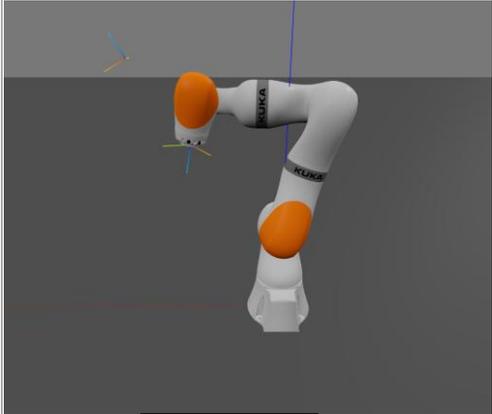 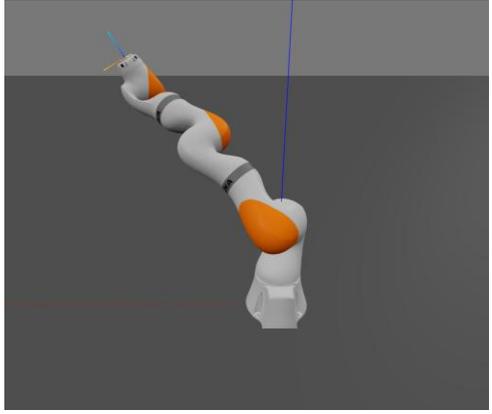 First Frame      Last Frame |
| Inspector | The final answer provided by RoboSolver matches the user's query. The LBR robot was successfully created and initialized in the ready joint configuration, and the position-based servoing was simulated using the specified end-effector transformation matrix in the world frame, with the correct proportional gain and velocity control profile. |

Table 7: Benchmark Test 3 – Query 3 and Corresponding Answers

| | |
|---|---|
| User | Create this robot for me:

The kinematic chain begins at a fixed base with a vertical offset of 0.1519 meters along the global Z-axis. It is followed by the first joint, which is revolute and rotates about the Z-axis, and then a link extending 0.1198 meters along the joint's local Y-axis.

Next comes the second joint, which rotates about the local Y-axis, followed by the second link, which has a compound offset: −0.0925 meters along the local Y-axis and 0.2437 meters along the local Z-axis — both as part of the same rigid link. The third joint follows, also rotating about the local Y-axis. Then comes the next link, which extends 0.2132 meters along the local Z-axis, followed by the fourth joint, rotating about the local Y-axis. After that, the next link extends 0.08505 meters along the local Y-axis, leading to the fifth joint, which rotates about the local Z-axis.

Finally, the next link extends 0.08535 meters along the local Z-axis, followed by the last revolute joint, which rotates about the local Y-axis. The end-effector is then positioned 0.0819 meters further along the local Y-axis. |
| RoboSolver | The custom robot has been successfully created with the specified kinematic chain. The robot has been assigned the ID 'd5a1c72a-6def-4c3f-82ee-9a817053225b'. Please use this ID for any further operations or queries related to this robot. |
| User | Then, solve the inverse kinematics of this robot for the following end-effector transformation matrix:

$$\begin{bmatrix} 0.36 & -0.79 & 0.5 & 0 \\ 0.7 & 0.58 & 0.41 & 0.23 \\ -0.61 & 0.2 & 0.76 & 0.57 \\ 0 & 0 & 0 & 1 \end{bmatrix}$$ |

| | Finally, consider the joint values obtained from the inverse kinematics as the pick-up joint configuration, and plot the robot's motion as it moves from the zero joint configuration to this pick-up configuration. |
|---|---|
| RoboSolver | The inverse kinematics for the robot with ID 'd5a1c72a-6def-4c3f-82ee-9a817053225b' was successfully solved for the given end-effector transformation matrix. The resulting joint configuration is [0.61, 0.98, -1.42, 1.01, 0.38, 0.14]. The robot's motion from the zero joint configuration to this pick-up configuration has been successfully plotted. 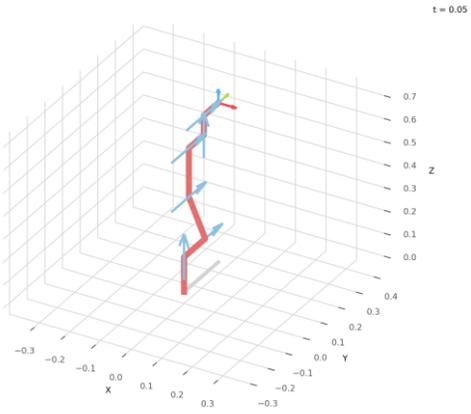 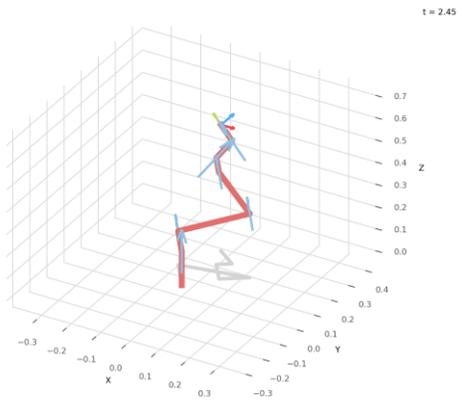<br>First Frame — Last Frame |

In this benchmark, the extractor pipeline responsible for image processing again relies on the Gemini 2.5 Pro Preview model, while the remaining agents and pipelines continue to operate based on GPT-4o. The results of this test, conducted on 10 queries—each evaluated three times—are presented in Tables 8 and 9. It is worth noting that this benchmark test is defined exclusively for the proposed framework, as answering certain queries requires robot control and simulation capabilities, which in turn demand an appropriate execution environment beyond the capabilities of standalone LLMs.

Table 8: RoboSolver Framework Performance Based on Evaluation Metrics in Benchmark Test 3.

| Model | Version | $\mathcal{M}_{Sup}$ | $\mathcal{M}_E$ | $\mathcal{M}_P$ | $\mathcal{M}_R$ | $\mathcal{M}_J$ | $\mathcal{M}_{SC}$ | $\mathcal{M}_C$ | $\mathcal{M}_T$ |
|---|---|---|---|---|---|---|---|---|---|
| GPT-4o & Gemini 2.5 Pro Preview | Framework | 1.00 | 1.00 | 0.95 | 0.95 | 1.00 | 0.83 | 0.97 | 0.96 |

Table 9: Average Token Usage, Execution Time, and Cost per Run for the RoboSolver Framework in Benchmark Test 3.

| Model | Version | Prompt Tokens | Completion Tokens | Total Tokens | Runtime (s) | Cost ($) |
|---|---|---|---|---|---|---|
| GPT-4o & Gemini 2.5 Pro Preview | Framework | 13490 | 2214 | 15704 | 48.31 | 0.045 |

According to the results presented in Table 8, the value of $\mathcal{M}_C$ for the framework is 0.97, indicating a very high level of accuracy in performing comprehensive tasks in the domain of robotic manipulators. On the other hand, the value of $\mathcal{M}_{SC}$, which is 0.83, shows that in 17% of the cases where the robosolver agent made an error, the inspector agent was not able to successfully correct it. Therefore, one possible direction for future development is improving the performance of the inspector agent to further enhance the $\mathcal{M}_C$ score. The average response time per run was approximately 48 seconds, which is acceptable. Additionally, the average execution cost was estimated to be around 5 cents, which may pose challenges for deploying this framework as a commercial product. However, by replacing GPT-4o with the DeepSeek-V3.2—and accepting a modest reduction in quality and an increase in response time—the framework could be offered to users as a practical product with a more affordable operating cost.

## 5 LIMITATIONS AND FUTURE WORKS

As previously stated, for image processing, it is assumed that the robot's visual input must be captured in a two-dimensional, planar view, ensuring that all robot components lie within a single plane. Furthermore, the user must provide the Joint Orientations in the World Frame to the framework to improve the accuracy of the image analysis performed by the VLM. These constraints indicate that VLMs have not yet reached a satisfactory level for spatial visualization understanding. Consequently, given that our framework relies on a VLM for image processing, a primary limitation of the proposed framework lies in its image-processing capabilities. This area can be addressed in future work to enhance the framework's generalizability in solving visual problems.

# 6 CONCLUSION

In this work, we introduced RoboSolver, an intelligent multi-agent framework that integrates LLMs, VLMs, and computational tools to support learning and problem-solving in robotics. Through three benchmark tests, the framework consistently outperformed the corresponding raw models across the defined accuracy metric. In forward-kinematics problems based on textual input data, the GPT-4o–based framework achieved an $\mathcal{M}_C$ of 0.97, representing a 67% improvement over the raw model. In comparison, the DeepSeek-V3.2–based variant attained a competitive $\mathcal{M}_C$ of 0.93, delivering acceptable performance while incurring substantially lower operational costs than the GPT-4o–based framework. For visually defined problems, the combination of GPT-4o with Gemini 2.5 Pro Preview yielded an $\mathcal{M}_C$ of 0.93, representing approximately a 20% gain over the raw VLM. In Benchmark 3, which included Jacobian computation, inverse kinematics, velocity and acceleration analysis, simulation, and robot control, the framework again achieved an $\mathcal{M}_C$ of 0.97, demonstrating robust performance across a wide range of robotic problems. The results indicate that equipping LLMs with dedicated computational tools and designing structured prompts that transform them into role-specialized agents significantly enhances their problem-solving capabilities. This integration enables accurate mathematical computation, structured reasoning, and reliable task execution beyond the limitations of standalone models. Moreover, within the multi-agent architecture, tasks are distributed among different agents in such a way that each agent remains focused on its specialized responsibility. This structured division of labor enhances accuracy, reduces error propagation, and improves coherence in the problem-solving process, ultimately leading to superior overall performance in complex robotic tasks. From an educational perspective, the framework can serve as an intelligent assistant that provides step-by-step solutions and visual simulations, thereby improving the quality of instruction and supporting deeper conceptual understanding. Finally, the RoboSolver framework can increase students' motivation while reducing their dependence on continuous human supervision by offering an interactive, automated problem-solving environment.

**Acknowledgment**

This work is based upon research funded by the Sharif University of Technology. We have used ChatGPT for improving the quality of the English writing of the paper.

**Statements & Declarations**

**Conflict of interest**

Author Alireza Taheri has received a research grant from the Sharif University of Technology. The authors Hamid Khabazi and Ali F. Meghdari declare that they have no conflict of interest.

**Availability of data and material (data transparency)**

All data from this project are available in the Social & Cognitive Robotics Laboratory archive.

**Code availability**

All of the codes are available in the archive of the Social & Cognitive Robotics Laboratory.

**Authors' contributions**

All authors contributed to the study's conception and design. Material preparation, data collection, and analysis were performed by Hamid Khabazi. Alireza Taheri and Ali F. Meghdari supervised this research. The first draft of the manuscript was written by Hamid Khabazi; and all authors commented on previous versions of the manuscript. All authors read and approved the final manuscript.

# A APPENDICES

Table 10: Benchmark Test 3 – Query 4 and Corresponding Answers

| | |
|---|---|
| User | Create this robot for me:<br><br>The robot's kinematic chain begins with a fixed translation of 0.333 meters along the Z-axis, followed by a revolute joint rotating about the same axis. Next comes another revolute joint around the local Y-axis. After that, the structure includes a fixed translation of 0.316 meters along the local Z-axis, then a revolute joint about the local Z-axis, and a fixed translation of 0.0825 meters along the local X-axis.<br><br>Subsequently, a flipped revolute joint around the local Y-axis is applied, followed by fixed translations of −0.0825 meters along the local X-axis and 0.384 meters along the local Z-axis. The chain then continues with a revolute joint about the local Z-axis, another flipped revolute joint about the local Y-axis, and a fixed translation of 0.088 meters along the local X-axis. Finally, the sequence includes a fixed rotation of $\pi$ radians about the local X-axis, a fixed translation of 0.107 meters along the local Z-axis, and ends with a revolute joint about the local Z-axis. |
| Supervisor | **\<Route to Extractor\>** |
| Extractor | Based on the provided description of the robot's kinematic chain, we can determine the elementary transforms (ETs) as follows:<br><br>1. **Fixed translation of 0.333 meters along the Z-axis**:<br>   - E1 = "tz(0.333)"<br>2. **Revolute joint rotating about the Z-axis**:<br>   - E2 = "Rz($\theta$1)"<br>3. **Revolute joint around the local Y-axis**:<br>   - E3 = "Ry($\theta$2)"<br>4. **Fixed translation of 0.316 meters along the local Z-axis**:<br>   - E4 = "tz(0.316)"<br>5. **Revolute joint about the local Z-axis**:<br>   - E5 = "Rz($\theta$3)"<br>6. **Fixed translation of 0.0825 meters along the local X-axis**:<br>   - E6 = "tx(0.0825)"<br>7. **Flipped revolute joint around the local Y-axis**:<br>   - E7 = "Ry(-$\theta$4)"<br>8. **Fixed translation of −0.0825 meters along the local X-axis**: |
| | - E8 = "tx(-0.0825)"<br>9. **Fixed translation of 0.384 meters along the local Z-axis**:<br>  - E9 = "tz(0.384)"<br>10. **Revolute joint about the local Z-axis**:<br>  - E10 = "Rz(θ5)"<br>11. **Flipped revolute joint about the local Y-axis**:<br>  - E11 = "Ry(-θ6)"<br>12. **Fixed translation of 0.088 meters along the local X-axis**:<br>  - E12 = "tx(0.088)"<br>13. **Fixed rotation of π radians about the local X-axis**:<br>  - E13 = "Rx(π)"<br>14. **Revolute joint about the local Z-axis**:<br>  - E14 = "Rz(θ7)"<br>15. **Fixed translation of 0.107 meters along the local Z-axis**:<br>  - E15 = "tz(0.107)"' |
| Planner | The user wants to create a robot based on a given kinematic chain described by a sequence of elementary transforms (ETs). To achieve this, we need to use the create_custom_userdefined_robot tool, which allows us to define a custom robot using a list of ET strings. This tool will take the robot name and the list of ETs as input and create the robot model accordingly. |
| RoboSolver | The robot has been successfully created based on the provided kinematic chain. The robot ID is fbc8fa66-180c-4b94-899a-bc75f411a4dc. Please use this ID for any further operations or queries related to this robot. |
| Inspector | The RoboSolver successfully created the robot based on the provided kinematic chain and provided a robot ID for further operations, which matches the user's request to create the robot. |
| User | Now compute the forward kinematics of the robot for the two given joint configurations: [-0.75, 1, -1, 0, 0, 1, 2] and [1, 2, -1, 1, 2, 3, 0].<br>Then, considering the first configuration as the initial state and the second as the final state, plot the robot's motion as it moves from the initial to the final configuration. |
| Supervisor | <Route to Planner> |
| Planner | The user wants to compute the forward kinematics for two different joint configurations and then plot the robot's motion from one configuration to the other. Since the robot has already been created with a specific kinematic chain, we can directly use the numerical tools for these tasks. First, we need to compute the forward kinematics for both joint |
| | configurations using the forward_kinematics_custom_userdefined_robot tool. After obtaining the forward kinematics, we will use the plot_custum_robot_motion tool to visualize the robot's motion from the initial to the final configuration. |
| RoboSolver | The forward kinematics for the robot with ID fbc8fa66-180c-4b94-899a-bc75f411a4dc have been computed for the given joint configurations. For the joint configuration ([-0.75, 1, -1, 0, 0, 1, 2]), the transformation matrix is: $$\begin{bmatrix} -0.77 & 0 & -0.64 & 0.39 \\ 0.31 & 0.87 & -0.38 & -0.52 \\ 0.56 & -0.48 & -0.67 & 0.66 \\ 0 & 0 & 0 & 1 \end{bmatrix}$$ For the joint configuration ([1, 2, -1, 1, 2, 3, 0]), the transformation matrix is: $$\begin{bmatrix} 0.85 & 0.4 & -0.35 & 0.09 \\ 0.26 & 0.27 & 0.93 & 0.65 \\ 0.46 & -0.88 & 0.12 & 0.34 \\ 0 & 0 & 0 & 1 \end{bmatrix}$$ dditionally, the robot's motion from the initial configuration ([-0.75, 1, -1, 0, 0, 1, 2]) to 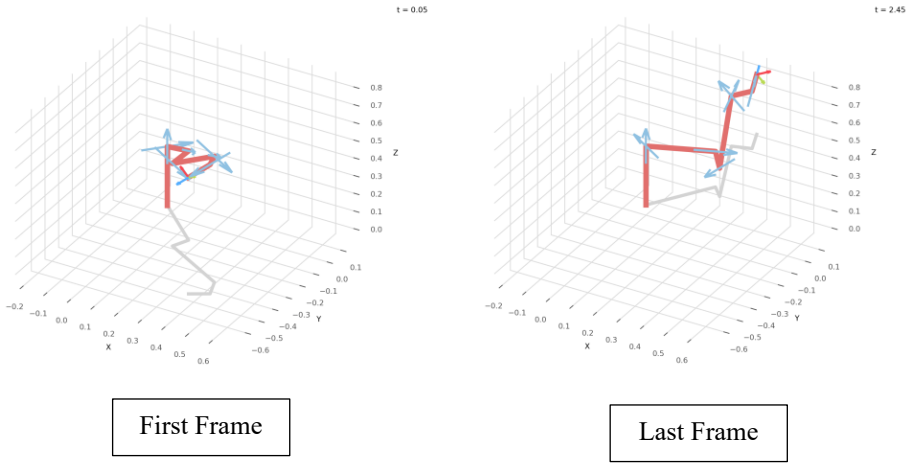 First Frame / Last Frame the final configuration ([1, 2, -1, 1, 2, 3, 0]) has been successfully plotted. |
| Inspector | The RoboSolver's final answer correctly computed the forward kinematics for the specified joint configurations and provided the transformation matrices. Additionally, it successfully plotted the robot's motion from the initial to the final configuration, as requested by the user. |

Table 11: Benchmark Test 3 – Query 5 and Corresponding Answers

| | |
|---|---|
| User | Solve the problem described in PDF file.<br><br>PDF Context:<br>Create a UR3 robot and plot its motion from the zero joint configuration to the joint configuration [-1, -1, -2.01, -1.58, -0.4, -1]. Then compute the Jacobian matrix in both the world frame and the end-effector frame. |
| Supervisor | <Route to Retriever> |
| Retriever | The problem description for solving the robotic problem involves creating a UR3 robot, plotting the motion of the robot from an initial joint value of zero to a joint configuration of [-1, -1, -2.01, -1.58, -0.4, -1], and computing the Jacobian matrix in both the world and end-effector frames. |
| Supervisor | <Route to Planner> |
| Planner | The user wants to create a UR3 robot, plot its motion from an initial joint configuration of zero to a specified joint configuration, and then compute the Jacobian matrix.<br>1. **Create the UR3 Robot**: The first step is to create the UR3 robot model using the predefined models available in the Robotics Toolbox library. This will provide us with a robot ID that will be used in subsequent steps.<br>2. **Plot Robot Motion**: Once the robot is created, we need to plot its motion from the zero joint configuration to the specified joint configuration. This requires the robot ID and the joint configurations.<br>3. **Compute Jacobian**: Finally, we need to compute the Jacobian matrix for the robot in both the world and end-effector frames. This also requires the robot ID and the joint configuration.<br>The tools needed in order are:<br>1. create_robotictoolbox_robot to create the UR3 robot.<br>2. plot_robot_motion to plot the motion of the robot.<br>3. compute_Jacobian to compute the Jacobian matrix. |
| RoboSolver | The UR3 robot has been successfully created with the ID d4ac73aa-4a75-4596-b502-d73b0f9139c7. the Jacobian matrix has been computed as follows:<br><br>**Jacobian in the World Frame:**<br><br>$$\begin{bmatrix} -0.24 & 0.15 & 0.04 & 0.02 & 0.02 & 0.0 \\ 0.07 & -0.23 & -0.06 & -0.04 & 0.02 & 0.0 \\ 0.0 & 0.16 & 0.29 & 0.08 & -0.07 & 0.0 \\ 0.0 & 0.84 & 0.84 & 0.84 & -0.54 & 0.8 \\ 0.0 & 0.54 & 0.54 & 0.54 & 0.84 & 0.46 \\ 1.0 & 0.0 & 0.0 & 0.0 & 0.12 & 0.39 \end{bmatrix}$$<br><br>**Jacobian in the End-Effector Frame:**<br><br>$$\begin{bmatrix} -0.16 & 0.08 & 0.12 & 0.03 & 0.0 & 0.0 \\ -0.18 & 0.12 & -0.12 & -0.02 & 0.04 & 0.0 \\ 0.07 & -0.29 & -0.25 & -0.08 & 0.07 & 0.0 \\ 0.39 & 0.92 & 0.92 & 0.92 & 0.0 & 1.0 \\ -0.6 & 0.21 & 0.21 & 0.21 & -0.84 & 0.0 \\ -0.7 & 0.33 & 0.33 & 0.33 & 0.54 & 0.0 \end{bmatrix}$$<br><br>Additionally, the motion of the robot from the zero joint configuration to the joint configuration of [-1, -1, -2.01, -1.58, -0.4, -1] has been plotted.<br><br>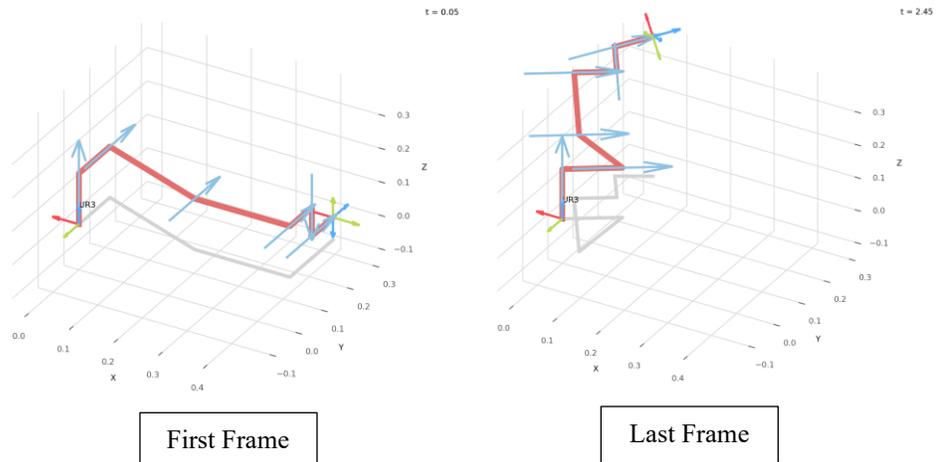<br><br>First Frame      Last Frame |
| Inspector | The RoboSolver's final answer matches the user's query. The UR3 robot was successfully created, the motion was plotted from the zero joint configuration to the specified joint |

configuration, and the Jacobian matrix was computed in both the world and end-effector frames, as requested by the user.

Table 12: Benchmark Test 3 – Query 6 and Corresponding Answers

| | |
|---|---|
| User | Compute the end-effector's velocity and acceleration based on the robot structure shown in the uploaded image.<br>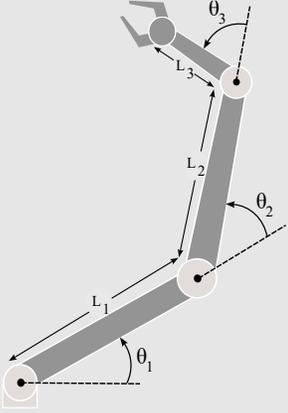<br>the joints orientation of the robot are:<br>joint 1: along the negative Y-axis<br>joint 2: along the negative Y-axis<br>joint 3: along the negative Y-axis. |
| Supervisor | <Route to Extractor> |
| Extractor | Based on the analysis of the robotic manipulator, we can determine the elementary transforms (ETs) for the system. Each joint and link contributes to the overall kinematics of the robot. Here is the sequence of ETs:<br>1. **Joint 1**: Revolute joint with rotation $\theta_1$ about the negative Y_U axis.<br>   - ET: "Ry(-θ1)"<br>2. **Link 1**: Translation along the X_U axis to reach Joint 2.<br>   - ET: "tx(L1)"<br>3. **Joint 2**: Revolute joint with rotation $\theta_2$ about the negative Y_U axis.<br>   - ET: "Ry(-θ2)"<br>4. **Link 2**: Translation along the X_U axis to reach Joint 3.<br>   - ET: "tx(L2)"<br>5. **Joint 3**: Revolute joint with rotation $\theta_3$ about the negative Y_U axis. |
| | - ET: "Ry(-θ3)"<br>6. **Link 3**: Translation along the X_U axis to reach the end effector.<br>  - ET: "tx(L3)" |
| Planner | The user wants to compute the end-effector's velocity and acceleration based on a symbolic robot structure defined by a sequence of elementary transforms (ETs). Since this is a symbolic problem, we need to use symbolic tools. First, we will compute the symbolic forward kinematics using the provided ETs. Then, we will compute the symbolic velocity and acceleration of the end-effector. The tools needed are:<br>1) Symbolic_Forward_Kinematic_ET to compute the forward kinematics,<br>2) Symbolic_EndEffector_Velocity to compute the velocity, and<br>3) Symbolic_EndEffector_Acceleration to compute the acceleration |
| RoboSolver | The symbolic forward kinematics, velocity, and acceleration of the end-effector for the given robot structure have been computed. The robot has been assigned the ID 12b71a2e-524d-41b6-9324-2b7c6849bcdd.<br><br>**Symbolic Forward Kinematics:**<br><br>$$\begin{bmatrix} \cos(\theta_1+\theta_2+\theta_3) & 0 & -\sin(\theta_1+\theta_2+\theta_3) & L_1\cos(\theta_1)+L_2\cos(\theta_1+\theta_2)+L_3\cos(\theta_1+\theta_2+\theta_3) \\ 0 & 1 & 0 & 0 \\ \sin(\theta_1+\theta_2+\theta_3) & 0 & \cos(\theta_1+\theta_2+\theta_3) & L_1\sin(\theta_1)+L_2\sin(\theta_1+\theta_2)+L_3\sin(\theta_1+\theta_2+\theta_3) \\ 0 & 0 & 0 & 1 \end{bmatrix}$$<br><br>**Symbolic End-Effector Velocity:**<br><br>$$\begin{bmatrix} -L_1\dot{\theta}_1 S_1 - L_2(\dot{\theta}_1+\dot{\theta}_2)S_{12} - L_3(\dot{\theta}_1+\dot{\theta}_2+\dot{\theta}_3)S_{123} \\ 0 \\ L_1\dot{\theta}_1 C_1 + L_2(\dot{\theta}_1+\dot{\theta}_2)C_{12} + L_3(\dot{\theta}_1+\dot{\theta}_2+\dot{\theta}_3)C_{123} \end{bmatrix}$$<br><br>**Symbolic End-Effector Acceleration:**<br><br>$$\begin{bmatrix} -L_1\ddot{\theta}_1 S_1 - L_2(\ddot{\theta}_1+\ddot{\theta}_2)S_{12} - L_3(\ddot{\theta}_1+\ddot{\theta}_2+\ddot{\theta}_3)S_{123} - L_1\dot{\theta}_1^2 C_1 - L_2(\dot{\theta}_1+\dot{\theta}_2)^2 C_{12} - L_3(\dot{\theta}_1+\dot{\theta}_2+\dot{\theta}_3)^2 C_{123} \\ 0 \\ -L_1 S_1 \dot{\theta}_1^2 + L_1 C_1 \ddot{\theta}_1 - L_2(\dot{\theta}_1+\dot{\theta}_2)^2 S_{12} + L_2(\ddot{\theta}_1+\ddot{\theta}_2)C_{12} - L_3(\dot{\theta}_1+\dot{\theta}_2+\dot{\theta}_3)^2 S_{123} + L_3(\ddot{\theta}_1+\ddot{\theta}_2+\ddot{\theta}_3)C_{123} \end{bmatrix}$$<br><br>These results provide the symbolic expressions for the kinematics of the robot based on the given elementary transforms. |
| Inspector | The RoboSolver's final answer provides the symbolic expressions for the forward kinematics, velocity, and acceleration of the end-effector, which matches the user's |

request to compute the end-effector's velocity and acceleration based on the robot structure.

Table 13: Benchmark Test 3 – Query 7 and Corresponding Answers

| | |
|---|---|
| User | Create a frankie robot and initialize it in the ready joint configuration. Then, with the following desired end-effector pose in the world frame: $$\begin{bmatrix} 0.77 & -0.47 & 0.42 & 0.5 \\ 0.42 & 0.87 & 0.22 & 0.5 \\ -0.47 & 0.0 & 0.87 & 0.75 \\ 0.0 & 0.0 & 0.0 & 1.0 \end{bmatrix}$$ Simulate the robot's motion from the ready joint configuration to the desired end-effector target using a position controller with a proportional gain of [3,3,3,3,3,3]. In this scenario, the target moves with a velocity of [0.15,0.15,0.05,0,0,0]. |
| Supervisor | <Route to Planner> |
| Planner | The user wants to simulate the position-based servoing of a predefined robot model, specifically the 'frankie' robot, using a desired end-effector position in the world frame. This requires the following steps: 1. **Create the Robot Model**: Use the 'create_robotictoolbox_robot' tool to create the 'frankie' robot model. This will provide us with the robot ID needed for further operations. 2. **Simulate Position-Based Servoing**: Use the 'simulate_robot_motion_Position_based_Servoing' tool to simulate the robot's motion. This tool requires the robot ID, the desired end-effector transformation matrix, the initial joint configuration ('qr' for ready state), the desired frame ('world'), the proportional gain, the velocity profile, and the dynamic target velocity. These steps will allow us to simulate the robot's behavior as per the user's request |
| RoboSolver | **Error occurred**: {'error': 'Invalid robot ID', 'current_tool_name': 'simulate_robot_motion_Position_based_Servoing', 'current_tool_arg': {'robot_id_local': 'frankie', 'end_effector_desired_transformation_matrix': [[0.77, -0.47, 0.42, 0.5], [0.42, 0.87, 0.22, 0.5], [-0.47, 0.0, 0.87, 0.75], [0.0, 0.0, 0.0, 1.0]], 'initial_joint_value': 'qr', |
| | 'desired_position_frame': 'world', 'proportional_gain': [3, 3, 3, 3, 3, 3], 'velocity_profile': True, 'dynamic_target': True, 'dynamic_target_velocity': [0.15, 0.15, 0.05, 0, 0, 0]} |
| Inspector | The error 'Invalid robot ID' occurred because the tool 'simulate_robot_motion_Position_based_Servoing' was called with an incorrect robot ID. The robot was created with the name 'frankie', but the tool requires the use of the generated Robot ID for further operations. The correct Robot ID 'a2b60db9-0497-4b0d-ad72-1b50c3aeaa9d' should be used instead of the name 'frankie'. Please update the tool call to use the correct Robot ID. |
| RoboSolver | The robot named 'frankie' was successfully created with the unique Robot ID 'a2b60db9-0497-4b0d-ad72-1b50c3aeaa9d'. The position servoing was simulated using the specified end-effector positions in the world frame, with a proportional gain of [3, 3, 3, 3, 3, 3] and a target position velocity of [0.15, 0.15, 0.05, 0, 0, 0]. The simulation was completed successfully. 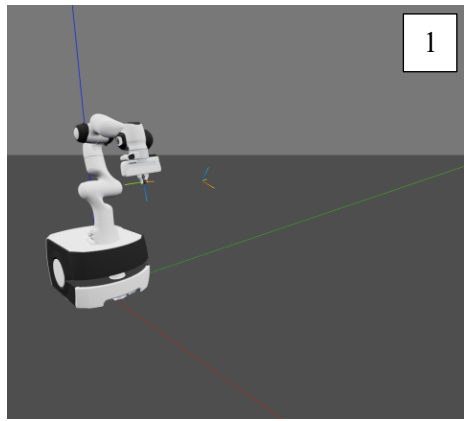 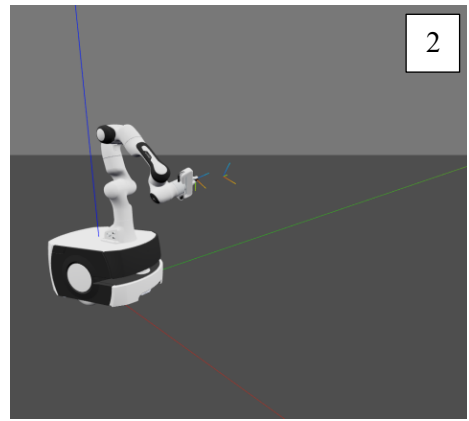 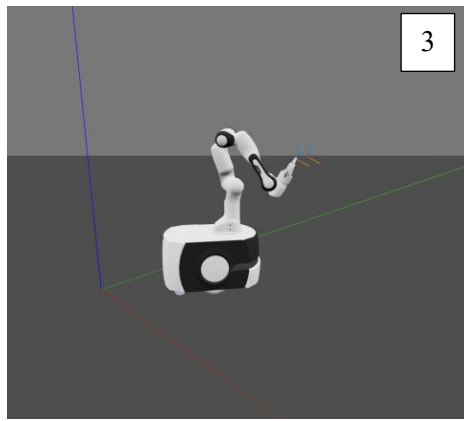 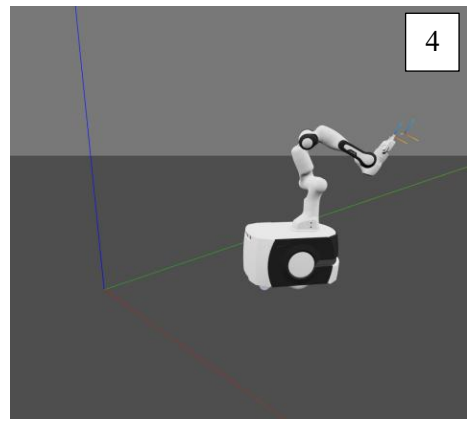 |
| Inspector | The RoboSolver's final answer matches the user's query. The robot 'frankie' was created and initialized as requested, and the position servoing was simulated using the specified end-effector positions, proportional gain, and target position velocity. The simulation was completed successfully, fulfilling all aspects of the user's request. |

Table 14: Benchmark Test 3 – Query 8 and Corresponding Answers

| | |
|---|---|
| User | Create the Panda robot and, starting from the ready initial joint configuration, simulate its motion with an end-effector velocity of $[0.04, 0, 0, 0, 0, 0.5]$ in the end-effector frame. Next, simulate the same task again, but this time in the world frame, using the same end-effector velocity. |
| RoboSolver | The Panda robot was successfully created with the ID 'fe22c220-17c0-41d6-8254-1fff2c70aedd'. The motion of the robot was simulated starting from the ready initial joint configuration. First, the simulation was performed with an end-effector velocity of [0.04, 0, 0, 0, 0, 0.5] in the end-effector frame. Then, the same task was simulated in the world frame with an end-effector velocity of [0.04, 0, 0, 0, 0, 0.5].<br><br>Simulation of the robot with an end-effector velocity of [0.04,0,0,0,0,0.5] in the end-effector frame:<br><br>First Frame   Last Frame<br>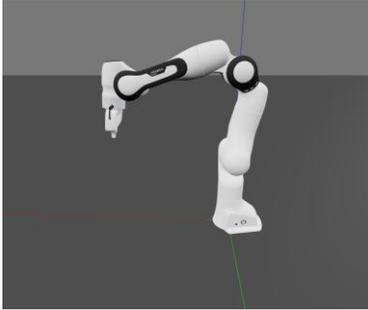 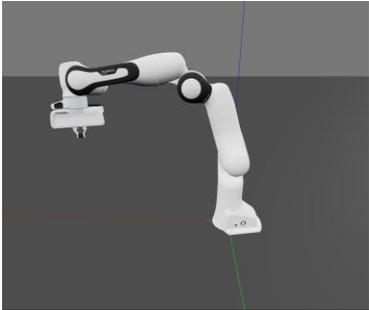<br>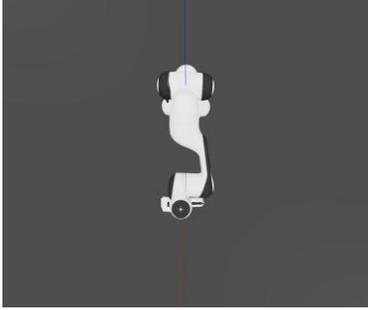 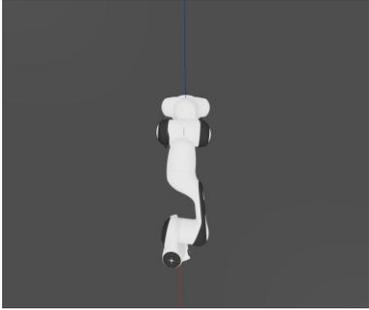 |

Simulation of the robot with an end-effector velocity of [0.04,0,0,0,0,0.5] in the world frame:

| First Frame | Last Frame |
|---|---|
| 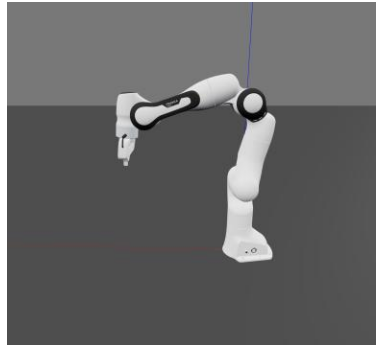 | 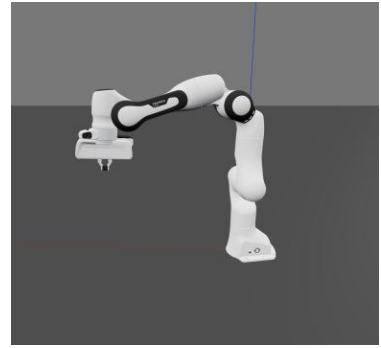 |
| 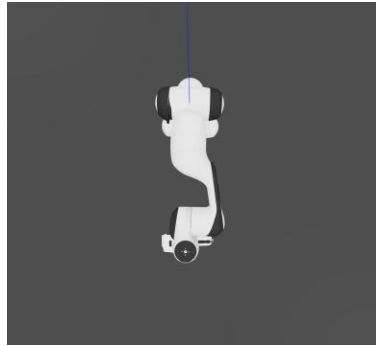 | 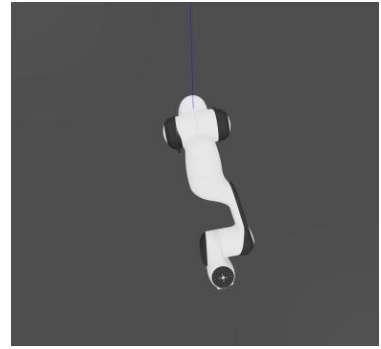 |